%% file: velocity_estimation_arxiv.tex
\documentclass[letterpaper, 10 pt, conference]{ieeeconf}  

\IEEEoverridecommandlockouts                              

\usepackage{graphics} 
\usepackage{graphicx}
\usepackage{mathptmx} 
\usepackage{times} 
\usepackage{amsmath} 
\usepackage{amssymb}  
\usepackage{siunitx}
\usepackage{blindtext}
\usepackage{booktabs}
\usepackage{xcolor}
\usepackage{color}
\usepackage{tikz}
\usetikzlibrary{positioning}
\usetikzlibrary{arrows}
\usetikzlibrary{calc}
\usetikzlibrary{shapes.multipart}
\usetikzlibrary{arrows.meta}
\usetikzlibrary{matrix}
\usetikzlibrary{fit}
\usetikzlibrary{decorations.pathreplacing}

\DeclareMathAlphabet{\mathcal}{OMS}{cmsy}{m}{n}

\title{\LARGE \bf
Self-Supervised Velocity Estimation for Automotive Radar Object Detection Networks
}

\author{Daniel Niederl\"ohner$^{1}$, Michael Ulrich$^{2}$, Sascha Braun$^{1}$, Daniel K\"ohler$^{1,3}$, Florian Faion$^{2}$, Claudius Gl\"aser$^{2}$, \\ Andr\'{e} Treptow$^{1}$, and Holger Blume$^{3}$
\thanks{$^{1}$Sascha Braun, Daniel K\"ohler, Andr\'{e} Treptow and Daniel Niederl\"ohner are with Robert Bosch GmbH, Cross-Domain Computing Solutions, Germany}%
\thanks{$^{2}$Michael Ulrich, Florian Faion and Claudius Gl\"aser are with Robert Bosch GmbH, Corporate Research, Germany 
	{\tt\small michael.ulrich2@bosch.com}}%
\thanks{$^{3}$Daniel K\"ohler and Holger Blume are with Leibniz University Hannover, Germany}%
}

\renewcommand{\baselinestretch}{1.02}

\begin{document}
	
\newcommand\copyrighttextinitial{%
	\scriptsize This work has been submitted to the IEEE for possible publication. Copyright may be transferred without notice, after which this version may no longer be accessible.}%

\newcommand\copyrighttextfinal{%
	\scriptsize\copyright\ 2022 IEEE. Personal use of this material is permitted. Permission from IEEE must be obtained for all other uses, in any current or future media, including reprinting/republishing this material for advertising or promotional purposes, creating new collective works, for resale or redistribution to servers or lists, or reuse of any copyrighted component of this work in other works.}%

\newcommand\copyrightnotice{%
	
	\begin{tikzpicture}[remember picture,overlay]%
	
	\node[anchor=south,yshift=10pt] at (current page.south) {{\parbox{\dimexpr\textwidth-\fboxsep-\fboxrule\relax}{\copyrighttextfinal}}};%
	\end{tikzpicture}%
	
}

\maketitle
\copyrightnotice%
\thispagestyle{empty}
\pagestyle{empty}

\begin{abstract}

This paper presents a method to learn the Cartesian velocity of objects using an object detection network on automotive radar data. 
The proposed method is self-supervised in terms of generating its own training signal for the velocities.
Labels are only required for single-frame, oriented bounding boxes (OBBs). 
Labels for the Cartesian velocities or contiguous sequences, which are expensive to obtain, are not required.
The general idea is to pre-train an object detection network without velocities using single-frame OBB labels, and then exploit the network's OBB predictions on unlabelled data for velocity training.
In detail, the network's OBB predictions of the unlabelled frames are updated to the timestamp of a labelled frame using the predicted velocities and the distances between the updated OBBs of the unlabelled frame and the OBB predictions of the labelled frame are used to generate a self-supervised training signal for the velocities.   
The detection network architecture is extended by a module to account for the temporal relation of multiple scans and a module to represent the radars' radial velocity measurements explicitly.
A two-step approach of first training only OBB detection, followed by training OBB detection and velocities is used.
Further, a pre-training with pseudo-labels generated from radar radial velocity measurements bootstraps the self-supervised method of this paper. 
Experiments on the publicly available nuScenes dataset show that the proposed method almost reaches the velocity estimation performance of a fully supervised training, but does not require expensive velocity labels.
Furthermore, we outperform a baseline method which uses only radial velocity measurements as labels. 

\end{abstract}

\section{Introduction}

One big challenge in the field of automated driving and advanced driver assistance systems is to perceive and understand the surrounding environment of a vehicle. For this purpose, it is necessary to reliably detect, track and classify all relevant objects. 
Automotive radar sensors are commonly used beside cameras, ultrasonic sensors and lidar systems. This is due to their low price, robustness in different weather conditions, and ability to measure radial velocities. 
Furthermore, radar sensors can serve as additional redundancy in highly automated driving.
Deep neural networks have shown promising results in solving the object detection tasks for various sensor modalities (e.g. \cite{Svenningsson2021, Liu2016, Zhou2017}).
Object detection networks are still relatively new for radar, whereas they can already be considered state-of-the-art for image processing \cite{Liu2016,He2017} and lidar \cite{Zhou2017,Yang2019}.
In recent research, radar data, mainly in the form of spectra or preprocessed point clouds, are used as input to a neural network to solve various problems such as semantic segmentation \cite{Schumann2020, Ouaknine_2021_ICCV}, classification \cite{Ulrich2021, Patel2019}, object detection \cite{Svenningsson2021, Wang2021a, Dreher2020, Xu2021, Danzer2019} or tracking \cite{Ebert2020, Tilly2020}.

To avoid confusion, we define the following terminology in this paper: 
\textit{Object detection} is used in terms of detecting oriented bounding boxes, as in the deep learning literature and not as \textit{detection} is used in the field of radar signal processing. 
Further, we use \textit{prediction} to describe the output of a neural network model, in contrast to how the term is used in tracking literature. 
\textit{Update} denotes the temporal transformation of the position of a bounding box through a motion model, similar to the term \textit{predict} in tracking literature. 

\begin{figure}[!tb]
	\centering
	\input{figures/teaser.tex}
	\caption{Velocity estimation in radar object detection networks is important for subsequent tracking. 	
	Labelling velocities is expensive, so this paper introduces a self-supervised framework using only oriented bounding box (OBB) labels.
	Radar point clouds at two timestamps $t_0$ and $t_0 - \Delta t$ (top row, the color of the points encodes the Doppler velocity) are fed to the network (middle row).
	The network's OBB proposals at both timestamps (bottom row) are used for training, where an OBB label exists only for the frame at $t_0$. 
	OBB proposals at timestamp $t_0 - \Delta t$ \protect\includegraphics[]{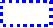} are first updated to timestamp $t_0$ using the network's velocity outputs and afterwards matched with the OBB proposals at timestamp $t_0$ \protect\includegraphics[]{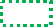} to generate a loss, which serves as a training signal for the predicted velocities. 
	By doing so, our network is able to output OBBs with velocity information without the need of explicit velocity labels.	
	}
	\label{fig:teaser}
	\vspace{-1.5em}
\end{figure}

\pagebreak
The combination of object detection and tracking is of great importance to obtain a stable environment model from the noisy and sparse radar measurements. 
Traditional methods typically create objects first and classify the detections afterwards.
Deep learning based detection networks perform both tasks simultaneously, overcoming the hand-crafted heuristics required to generate individual objects from multiple radar reflections.
These networks usually consume a single scan or multiple time-aggregated scans from one or multiple radar sensors and provide a set of oriented bounding boxes (OBBs) that contain attributes such as class, position, extent, and orientation for all relevant objects in a specific scene. 
Afterwards, the detections can be tracked for example by using a recursive Bayesian state estimation (e.g. a Kalman Filter) or a tracking network \cite{Tilly2020, Yin2021, Luo2020} on top of the detector. 

The velocity of the objects is of great interest for tracking. 
Radar sensors can only measure a radial projection of the object's velocity, i.e. tangential movements to the sensor are not detected.
However, the full Cartesian velocity estimates could improve the tracking performance or facilitate track initialization, for instance. 
Reliable velocity labels are difficult to obtain, in comparison to the above mentioned box attributes because the datasets are usually labelled by humans using lidar and camera data, which do not provide direct velocity information. 
A common solution is to label the objects position in sequences and estimate the velocity from consecutive frames \cite{Tilly2020, Yang2020}. 
Nevertheless, labelling sequences is significantly more expensive than labelling individual frames because many more scenes are required to achieve comparable diversity in the dataset.
In this paper, we propose a solution to this problem with the following contributions:
\begin{itemize}
	\item We propose a radar object detection network that uses point clouds and outputs not only the class, position, extent, and orientation, but also the velocity of objects on a single-shot basis. 
	This network is based on a state-of-the-art architecture \cite{Yang2019}, for which we introduce extensions to better account for temporal relations and the radar radial velocity measurements. 
	\item We introduce a self-supervised method to learn the full Cartesian velocities.
	Compared to the Doppler measurement of the radar, this also includes tangential movements to the sensor. 
	Our approach needs only the usual OBB labels to train the detection. 
	Labels for the Cartesian velocity are not required, which reduces the labelling effort significantly. 
	\item We benchmark our algorithms on the publicly available nuScenes dataset \cite{Caesar2020} and show that we achieve comparable results to a fully supervised velocity regression.  
\end{itemize}

\section{Related Work}
\subsection{Radar Object Detection}
Common approaches often use classical clustering and/or tracking methods to first create object tracks, which are subsequently classified \cite{Ulrich2021, Schumann2017, Schumann2018b, Scheiner2020}.
In contrast, object detection based approaches simultaneously locate and classify objects using deep neural networks.
Such object detection networks can be distinguished by the type of radar input data they consume. 
Spectrum-based models for object detection utilize 3D radar spectra (range, velocity, azimuth) \cite{Palffy2020} or reduced 2D \cite{Brodeski2019,Wang2021a, Meyer2021} or 1D \cite{Ulrich2018a,Ulrich2018b} projections.

Other models process radar data in the more abstract form of point clouds, which consist of a list of reflections enhanced with additional features, such as radial velocity~$v_r$ and radar cross section~$\sigma$ (RCS). These point cloud based networks can be further differentiated into grid-based and point-based architectures. Grid-based approaches first render the point cloud into a 2D bird eye view (BEV) or 3D voxel grid using hand-crafted operations \cite{Dreher2020, Nobis2021b, Lee2020, Meyer2019a,Scheiner2021} or learned feature-encoders \cite{Lang2019, Xu2021,Scheiner2021} and subsequently apply convolutional backbones to the grid.

In contrast, point-based methods directly extract features from the radar point cloud without an intermediate grid rendering step. For example \cite{Danzer2019, Bansal2020, Schumann2020} adopt PointNet/PointNet++ to achieve a hierarchical feature aggregation of neighbouring points. Other methods in the literature use a graph neural network (GNN) \cite{Svenningsson2021}, kernel point convolutions \cite{Nobis2021a} or an attention mechanism \cite{Bai2021} to exchange contextual information between points.

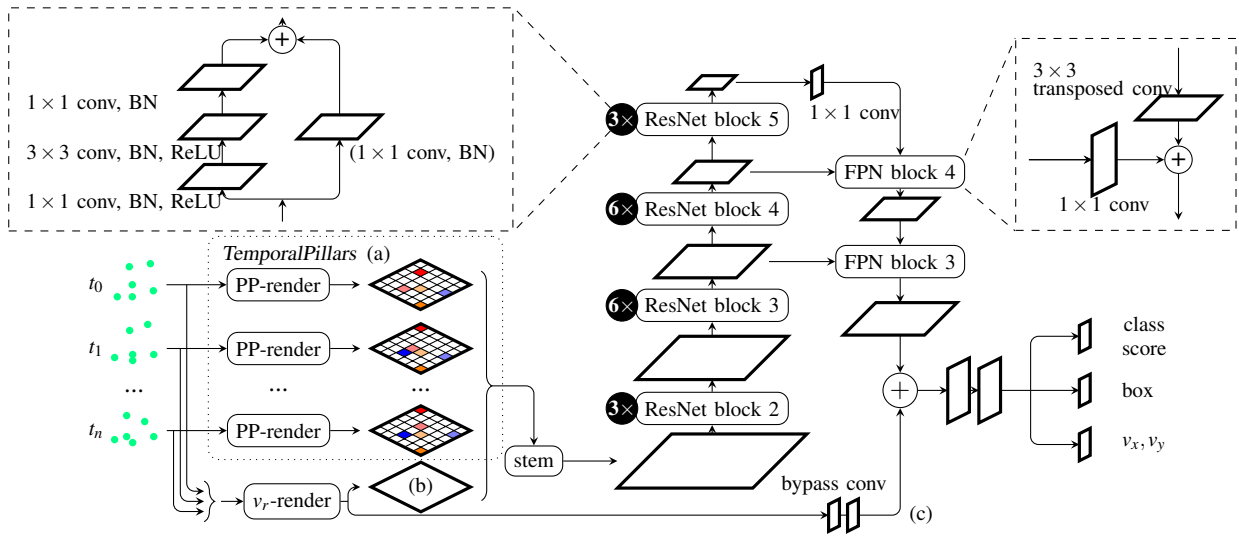
\begin{figure*}[!tb]
	\centering
	\input{figures/network_architecture_vel.tex}
	\caption{Architecture of the proposed model. 
		We added: (a) a module \textsl{TemporalPillars} to stack feature maps of multiple scans, which are individually processed similar to the PointPillars \cite{Lang2019} (PP) rendering, (b) a dedicated feature map for radial velocity measurements ($v_r$-map) and (c) a shortcut connection of the $v_r$-map to the detection head. The backbone and the detection head are analogous to \cite{Yang2019}, with an additional output for the velocity in Cartesian coordinates. $n\times n$ (transposed) conv refers to a 2D (transposed) convolution with kernel size $n\times n$. BN refers to batch normalization.}
	\label{fig:network_architecture}
\end{figure*}

\subsection{Velocity Estimation}
Radar sensors can only measure a radial projection of the object's velocity which does not contain tangential motion information. 
Yet, it is possible to calculate the full motion of rigid bodies instantaneously in many cases, using multiple Doppler and azimuth angle measurements. 
For example, the authors of \cite{Kellner2013, Kellner2014} use the Doppler over azimuth angle profile of multiple reflections on the same object and a RANSAC algorithm to fit the Cartesian velocity. 
One disadvantage is that these methods need a handcrafted association of points to objects. 
Deep learning based methods can overcome this bottleneck by learning the association. 
However, there are only few publications on this topic so far.
Often a sensor fusion network is used, that predicts the objects Cartesian velocity components $\left(v_x, v_y\right) $ in a vehicle coordinate system as additional box regression attributes. 
For example, \cite{Nabati2021} fuses radar and camera data in a CenterNet, while \cite{Yang2020} proposes RadarNet, a radar and lidar fusion network which exploits an additional attention-based late fusion step to refine the predicted velocities. 
With \cite{Svenningsson2021} also a radar-only approach has been published. 
The authors use a GNN for object detection and additionally regress the speed (absolute value of the velocity), assuming that the velocity direction coincides with the box orientation.

Another common method is to use a joint detection and tracking network that processes multiple sequential frames and stores the tracking history internally. The velocities are first predicted by the detection part and filtered afterwards by the tracking network. 
Such approaches are mostly applied to lidar (e.g. \cite{Yin2021, Luo2020}), but also a first radar-only network has been published \cite{Tilly2020}.

All above approaches require labelled velocities to perform a supervised training.
In contrast, the proposed method of this paper learns velocities implicitly and does not require velocity labels. 

\section{Method}
\subsection{Overview}
This work proposes a radar-only object detection network that outputs 3D bounding boxes with additional Cartesian velocity information without the need of sequence or velocity labels in a self-supervised manner. 
In this work, self-supervision refers only to the velocity learning and means that all training signals to improve velocity predictions are generated by the network itself on the fly and no explicit velocity or sequence labels are needed.
OBB labels are still required, but labels of single frames (non-contiguous time steps) are sufficient.

The core idea is to split the training into two steps, similar to \cite{Yin2021}. During the first step (here called \textit{detection step}), the network is trained in a fully supervised way to predict OBBs using 3D box labels and the corresponding radar input data. The velocity output is ignored. In the second step (here called \textit{velocity step}) the network is fed with an unlabelled input, which is close in time to the first step, cf. Fig.\,\ref{fig:teaser}.
The position of the OBB detections are then updated to the timestamp of the \textit{detection step} using the predicted velocities and a fixed (not trainable) constant velocity model. Afterwards, highly confident OBBs of both steps are matched based on the minimal Euclidean distance of the box centers (i.e. OBBs closest to each other are matched). 
Finally, the distances between the centers of matched boxes are used as a self-supervised loss during backpropagation to train the velocity output.
The constant velocity model is an appropriate choice, as long as the time difference is sufficiently small compared to the objects' dynamics. 
The two-step approach only affects the network training. 
During inference, the OBBs and velocities are predicted in a single step. 

\subsection{Base Network Architecture}
We use a grid-based architecture as depicted in Fig.\,\ref{fig:network_architecture} that processes radar point clouds. First, the reflections are projected onto a BEV grid using a PointPillars-like grid renderer, see \cite{Lang2019}. Similar to \cite{Scheiner2021} we found that the grid resolution should be rather coarse (\SI{0.5}{m}$\times$\SI{0.5}{m}), due to the sparsity of radar data.
Furthermore, consistently with \cite{Scheiner2021}, we found empirically that the backbone and head used in \cite{Lang2019} does not perform well with automotive radar data. 
Instead we adopt the lightweight approach from \cite{Yang2019} that uses a 2D convolutional backbone consisting of a residual network \cite{He2016} and a feature pyramid network \cite{Lin2017} to process the BEV feature maps at different resolutions. 
For a detailed description we refer to \cite{Yang2019}.
A multitask head predicts both the class scores and the bounding box parameters. 
In this work we only train the class \textit{car}. 
Theoretically, other classes can be included in the architecture, but we did not investigate the implications further. 

\begin{figure}[!tb]
	\centering
	\input{figures/matching.tex}
	\caption{Scheme of the training process. The networks OBB predictions at two timestamps $t_0$ and $t_0 - \Delta t$ are used for velocity training. Box positions at timestamp $t_0 - \Delta t$ are updated to $t_0$ using the predicted velocities and highly confident proposals are matched with boxes from $t_0$ to obtain a self-supervised loss $\mathcal{L}_{\text{vel}}$. OBB labels at timestamp $t_0$ are used to calculate the supervised detection losses $\mathcal{L}_{\text{cls}}$ and $\mathcal{L}_{\text{box}}$. }
	\label{fig:matching}
\end{figure}
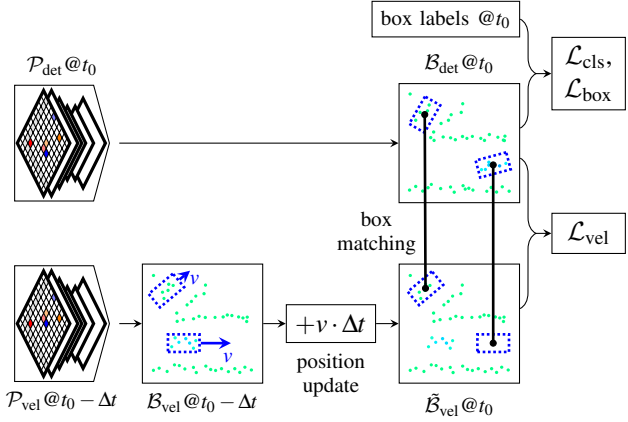

\subsection{Velocity Extension}
We added three velocity specific extensions, in addition to the baseline network architecture. 
We believe that these extensions are easily applicable to a wide range of other radar object detection network architectures:
\label{sec:velocity_extension}

\textbf{Velocity output} The multitask head is extended with a third output for velocity estimation. This output regresses the Cartesian velocities over ground $\left(v_x, v_y\right)$ in the ego vehicle coordinate system by adding another 2D convolutional layer with kernel size $3\times3$ and linear activation. 
The output directly predicts the $v_x$ and $v_y$ in meters per second without further encoding.

\textbf{Velocity shortcut} In addition, we found that adding the ego motion compensated radial Doppler velocity $v_r$ (without further transformations) to high level feature maps slightly improves the velocity regression. 
Therefore, we create an additional BEV map ($v_r$-map) with the same resolution as the PointPillars map, holding the maximum $v_r$ of all radar points within a cell. 
Empty cells are padded with $0$. 
This $v_r$-map is concatenated to the output of the PointPillars-like feature maps before the grid is passed to the backbone, cf. Fig.~\ref{fig:network_architecture}(b). 
Furthermore, there is a shortcut (bypass) path for the radial Doppler information ($v_r$-map) into high level feature maps, similar to \cite{Yang2020}. 
On this shortcut path, the $v_r$-map is clipped at $\pm$\SI{50}{m/s} and normalized. 
Afterwards, we downsample the $v_r$-map to the same resolution as the final feature map by applying a 2D convolution with kernel size $3\times3$ and $16$ channels and a bottleneck 2D convolution with kernel size $1\times1$ and $1$ channel, both followed by a max-pooling operation with stride $2$. 
This feature map is then added cell-wise to the feature maps before the head.

\textbf{TemporalPillars} Another change we introduced is a modification to the PointPillars renderer \cite{Lang2019}. 
Radar point clouds are typically quite sparse. 
It is common practice to increase the density by aggregating multiple consecutive scans \cite{Svenningsson2021, WangLeichen2020}. 
Previous methods from the literature simply used the entire aggregated point cloud $\mathcal{P} = \{ \mathcal{P}_1,\dots, \mathcal{P}_n\}$, where $n$ denotes the number of input scans used.
In contrast, we feed the point cloud of each scan $\mathcal{P}_i$ (holding all radar sensors) individually into a separate PointPillars renderer and then concatenate all $n$ output maps, to better account for temporal information.
The ego vehicle movement is compensated beforehand, such that all scans $\mathcal{P}_i$ refer to a common timestamp. 
Note that the motion compensation only works correctly for stationary targets, which is why we add the time difference between the original timestamp of a point and the compensated timestamp as an additional input feature to the network, cf. Sec.\,\ref{sec:detection_step}.
The proposed TemporalPillars is equivalent to the PointPillars renderer for the special case of $n=1$. 

\subsection{Training of the Network}

The network is trained using a combination of two different steps: First the fully supervised \textit{detection step} and second the self-supervised \textit{velocity step}. 
The \textit{velocity step} does not require any labels (self-supervised), while the usual single-snapshot OBB labels are sufficient for the \textit{detection step} (no labelled Cartesian velocity).
In both steps, the format of the input point cloud and the network architecture are the same, only training procedure differs.
A general overview of the training is shown in Fig.\,\ref{fig:matching}.

\textbf{Detection step}
\label{sec:detection_step}
In this step, only the box regression and classification part without velocity output is trained in a traditional supervised way with 3D box labels and the corresponding radar input point cloud $\mathcal{P}_{\text{det}} = \{ p_1,\dots, {p}_m\}$. 
The point cloud is a set of $m$ unordered radar reflections $p_i = \{\underline{x}_{p,i}, v_{r,i}, \sigma_i, \alpha_i, \Delta t_i \}$ from all radar sensors from one or multiple temporal aggregated scans, where $\underline{x}_{p,i} \in \mathbb{R}^3$ denotes the ego movement compensated 3D Cartesian position in a common vehicle coordinate system, $v_{r,i}\in  \mathbb{R}$  the ego motion compensated radial Doppler velocity, $\sigma_i \in  \mathbb{R}$ the radar cross section in dBsm, $\alpha_i \in  \mathbb{R}$ the azimuth angle in sensor coordinates and $\Delta t_i \in  \mathbb{R}$ the time difference of the measurement to a common reference timestamp of $\mathcal{P}_{\text{det}}$, which is the timestamp of box labels. 

The network predicts a set of three dimensional OBB proposals $\mathcal{B}_{\text{det}}$. 
One OBB $b_i = \{ \underline{x}_{c,i}, l_i, w_i, h_i, \theta_i, v_i\}$ is composed of a center
position $\underline{x}_{c,i}  \in \mathbb{R}^3 $ in vehicle coordinates, a length $l_i \in \mathbb{R}$, a width $w_i \in \mathbb{R}$, a height $h_i \in \mathbb{R}$, a yaw orientation angle $\theta_i \in \mathbb{R}$ and a Cartesian velocity vector $v_i = [v_{x,i}, v_{y,i}]^\top\in \mathbb{R}^2 $ in vehicle coordinates. The roll and pitch angles as well as the velocity in $z$-direction are omitted. 
We use the same box encoding as described in \cite{Yang2019} for training and decode the box-codes during inference. The velocity output is not encoded and ignored during the \textit{detection step}. 
For classification we use a focal loss $\mathcal{L}_\text{cls}$ on all grid cells and for box regression a smooth $L1$ loss $\mathcal{L}_\text{box}$ only on positive cells (i.e. cells close to a ground truth box). 
The overall detection loss is composed as a multitask loss $\mathcal{L}_{\text{det}} = c_\text{box}  \mathcal{L}_\text{box} + c_\text{cls}  \mathcal{L}_\text{cls}$ with the constant weight factors $c_\text{box}$ and $c_\text{cls}$. 

\textbf{Velocity step}
In the \textit{velocity step}, the network uses an unlabelled point cloud $\mathcal{P}_\text{vel}$ from a previous frame.  
We found that a fixed time offset $\Delta t_{\text{vel} \rightarrow \text{det}} \approx 600$ \si{ms} is a suitable value.
The point cloud is compensated to the same reference timestamp for both steps (timestamp of $\mathcal{P}_{\text{det}}$), using the movement of the ego vehicle.
This means that static points will be located at the same position in both steps. 
The deviation of the position of dynamic objects is intended to train the velocity predictions, as described in the sequel. 

In the forward path, the network produces a set of OBB proposals $\mathcal{B}_\text{vel}$ from $\mathcal{P}_\text{vel}$.
The velocity output is trained as follows:
First, the proposals are updated to the reference timestamp of $\mathcal{P}_{\text{det}}$ by updating the center positions of all boxes in $\mathcal{B}_\text{vel}$ using a constant velocity model
\begin{align}
	\tilde{\underline{x}}_{c,i} = \underline{x}_{c,i} + [v_{x,i}\Delta t_{\text{vel} \rightarrow \text{det}}, v_{y,i}\Delta t_{\text{vel} \rightarrow \text{det}}, 0]^\top \label{eq:box_update}
\end{align}
while all other box parameters remain unchanged. The updated set of box proposals $\tilde{\mathcal{B}}_\text{vel}$ now refers to the same timestamp as the network output from the \textit{detection step}. 
We keep only highly confident boxes and remove all proposals with a high background classification softmax score $s_{i, \text{bg}}$ of the corresponding box proposal $b_i$ for both \textit{detection step} and updated \textit{velocity step}
\begin{align}
\mathcal{\tilde{B}}_{\text{vel},\text{conf}} =& \{b_i \in \tilde{\mathcal{B}}_\text{vel} \mid s_{i, \text{bg}} < \epsilon_\text{conf} \}\label{eq:box_vel_conf}\\
\mathcal{B}_{\text{det},\text{conf}} =& \{b_i \in \mathcal{B}_\text{det} \mid s_{i, \text{bg}} < \epsilon_\text{conf} \}\label{eq:box_det_conf}
\end{align}
where $\epsilon_\text{conf}$ is a constant threshold parameter.
A suitable value was derived from the score distributions of all true-positive and false-positive predictions in the dataset, such that there are enough true-positive examples for training while keeping the amount of false-positives sufficiently small.
Finally, we match the predictions of both steps by calculating the Euclidean distance $d_{i,j}$ of centers for all box pairs $(b_i,b_j)$ between $\mathcal{\tilde{B}}_{\text{vel},\text{conf}}$ and $\mathcal{B}_{\text{det},\text{conf}}$.
The boxes are matched according to the smallest distances and the center distances $d_{i,j}$ of all matches $\mathcal{M}$ are summed to obtain the self-supervised velocity loss 

\begin{align}
\mathcal{L}_\text{vel} = \frac{c_\text{vel}}{\lvert \mathcal{M} \rvert} \sum_{(i,j) \in \mathcal{M}} d_{i,j}
\end{align}
where $c_\text{vel}$ denotes a constant weight factor and ${\lvert \mathcal{M} \rvert}$ the number of matches. 
If the cardinality of $\mathcal{\tilde{B}}_{\text{vel},\text{conf}}$ and $\mathcal{B}_{\text{det},\text{conf}}$ is not equal, we only use $m = \min (\lvert \mathcal{\tilde{B}}_{\text{vel},\text{conf}} \rvert, \lvert \mathcal{B}_{\text{det},\text{conf}} \rvert)$ matches with the smallest Euclidean distance for loss calculation. The velocity output is trained by applying backpropagation to $\mathcal{L}_\text{vel}$.
We want to highlight, that this step does not incorporate any manual labels.

\textbf{Training procedure}
\label{sec:training_phases}
The network is trained in two phases. 
During the first phase we only apply the supervised \textit{detection step} for several epochs to achieve good detection results. 
The velocity output is ignored. 
In the second phase, both steps are performed alternately. 
Furthermore, our experiments showed that the self-supervised step can be stabilized if the velocity output is pre-trained in the first phase. 
For this purpose, we use the maximum measured radial Doppler velocity within a ground truth box and project the value to the direction of the box heading. 
The resulting pseudo labels $(v_{r,x}, v_{r,y})$ are used to train the velocity output by calculating an additional supervised smooth $L1$ loss $\mathcal{L}_{\text{vr}}$ which is added to the multitask loss $\mathcal{L}_{\text{det},\text{vr}} = \mathcal{L}_\text{det} + c_{\text{vr}} \mathcal{L}_{\text{vr}}$ with a constant weight factor $c_{\text{vr}}$. Note that $v_r$ is only a radial projection of the objects velocity, so the second training phase is still necessary to train the full Cartesian velocity.  $(v_{r,x}, v_{r,y})$ are not used during the second training phase. 

\section{Experiments}
We conduct our experiments on the publicly available nuScenes dataset \cite{Caesar2020}, which consists of urban real world scenarios recorded in different countries.
The evaluation uses the official nuScenes object detection metrics \cite{Caesar2020} and does not perform radar-specific adjustments, such as removing static objects, filtering the radar field of view, or ignoring boxes without radar points. 
This makes our results more comparable, even with other modalities. 
In this work we focus on the \textit{average velocity error} (AVE) to evaluate the quality of the velocity predictions.
The AVE is a true-positive metric, meaning that only boxes that match the ground truth are included in the calculation, with a distance threshold of two meters. False-positive detections are not considered in the calculation.
It describes the average velocity error between predictions and ground truth boxes in meters per second, see \cite{Caesar2020}.
Further, the detection performance is evaluated in terms of \textit{average precision} at a distance threshold of four meters (AP4.0) and the mean\textit{ average precision} of all distance thresholds $\mathbb{D}=\{0.5, 1, 2, 4\}$ meters (AP) per class. 
We refer to  \cite{Caesar2020} for a more detailed explanation of the metrics.

The hyperparameter were determined empirically. The model is trained for $15$ epochs in each of the two training phases using an initial learning rate of $1 \times 10^{-3}$ in the first phase and  $0.5 \times 10^{-3}$ in the second phase. 
The constants are set to  $c_\text{cls} = 10$, $c_\text{box} = 0.5$, $c_{\text{vr}}=0.1$, $c_\text{vel} = 0.05$ and $\epsilon_\text{conf} = 0.5$. 
We always use all five radar sensors and augment the input data during training by randomly rotating the entire point cloud in a range of $\pm5$ degree. 
Seven consecutive aggregated scans ($\widehat{=}450$ \si{ms} for radar sensors in the nuScenes dataset) of radar data are used per frame.
All evaluations are preformed on the official nuScenes validation split.
The ground truth values for the velocity are derived based on the position change in sequential frames. 
We only train on class \textit{car}.
Other classes are possible, but for radar the detection performance is typically worse, due to the sparsity of the point cloud and the class imbalance in the dataset. 
This might affect the self-supervised \textit{velocity step} negatively.

For lack of previous published work, we conduct a supervised training of velocities as an error lower bound. 
Furthermore, we compare our method against a supervised training that only uses the maximum measured radial Doppler velocity within a ground truth box projected to the direction of the box heading $(v_{r,x}, v_{r,y})$ as labels. 
Both supervised approaches are trained for $15$ epochs without applying the \textit{velocity step} (i.e. only the first training phase is used).

\newcommand{\picturewithcoordinates}[5]{
	\begin{tikzpicture}
	\node (image) at (0,0) {\includegraphics[height=0.1\paperheight]{#1}};
	\draw [red,-{Stealth}, line width=1pt] #2--#3; 
	\draw [blue,-{Stealth}, line width=1pt] #4--#5; 
	\end{tikzpicture}
}

\begin{figure*}[!tb]
	\centering
	\begin{tabular}{c c c c}
		& \textbf{Camera} & \textbf{Self-supervised (ours)} & \textbf{Projected Doppler}\\
		\raisebox{0.85\totalheight}{\rotatebox[origin=c]{90}{Example $1$}}
		& \picturewithcoordinates{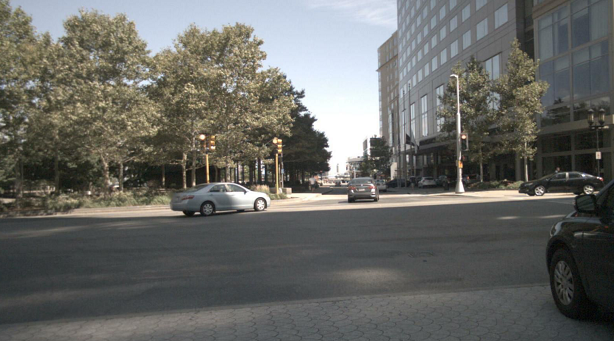}{(0, -1.35)}{(0, -0.85)}{(0, -1.35)}{(-1.0, -1.35)}
		& \picturewithcoordinates{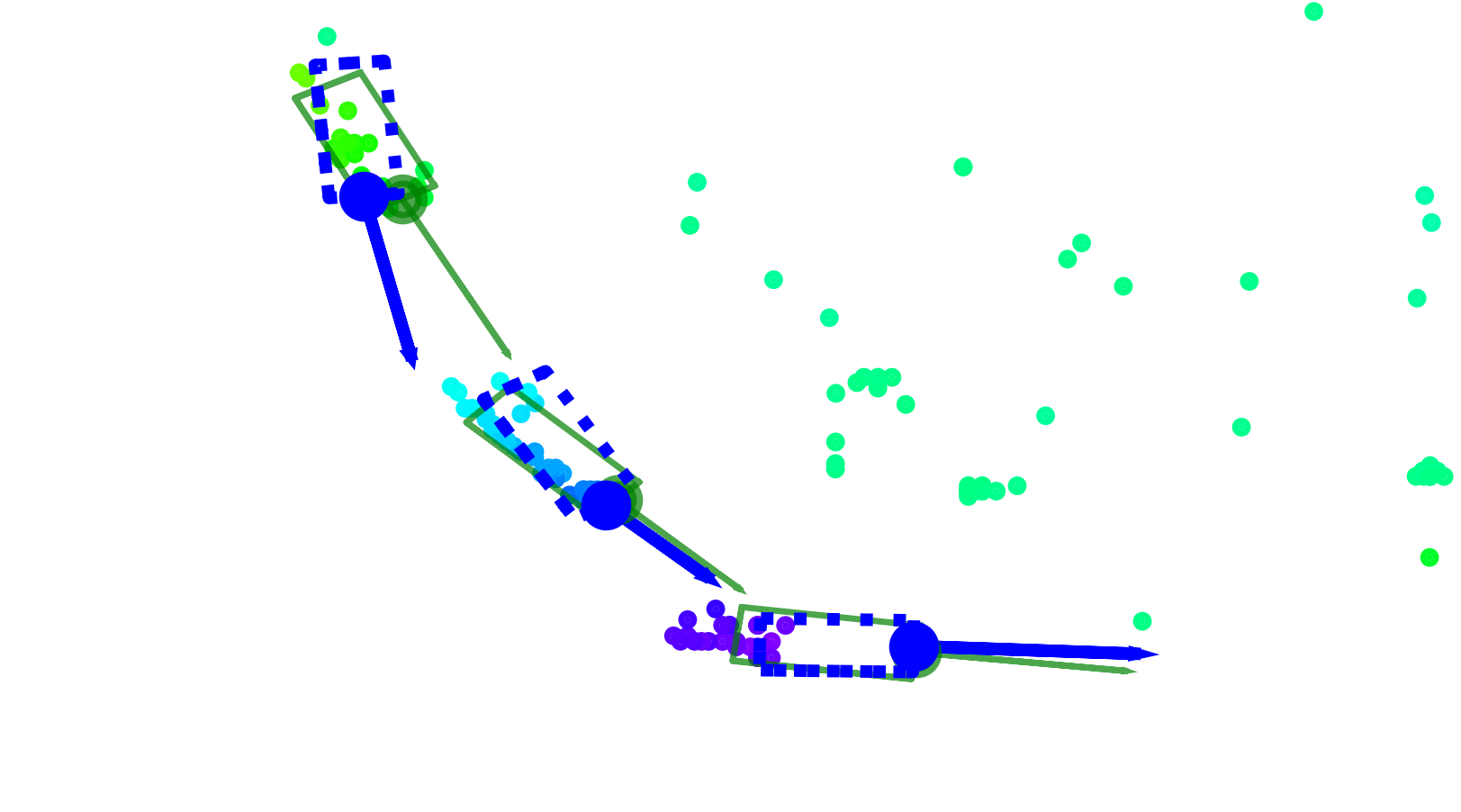}{(-2.5, 0)}{(-2.0, 0)}{(-2.5, 0)}{(-2.5, 1.0)}
		& \picturewithcoordinates{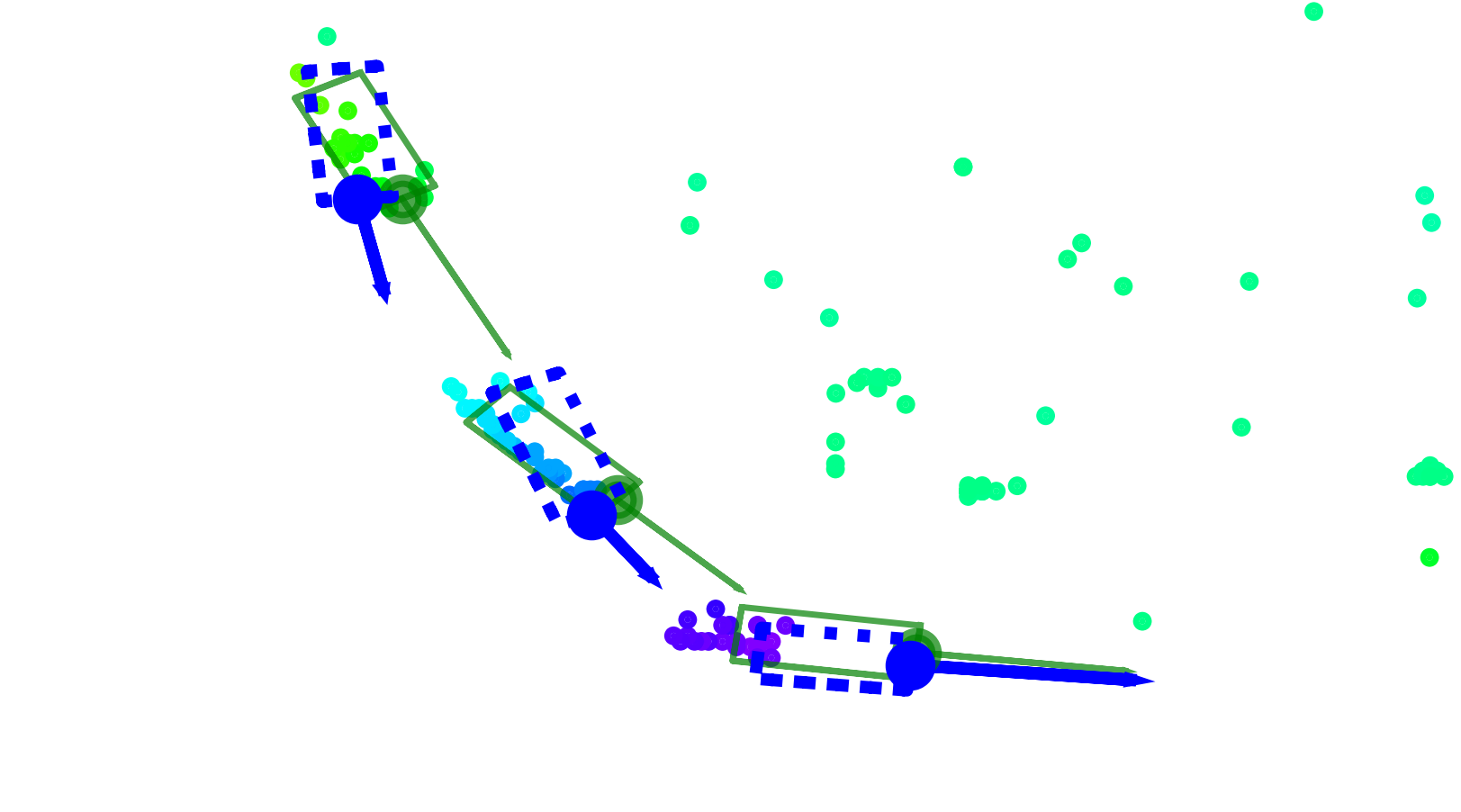}{(-2.5, 0)}{(-2.0, 0)}{(-2.5, 0)}{(-2.5, 1.0)}
		\\ \hline
		\raisebox{0.85\totalheight}{\rotatebox[origin=c]{90}{Example $2$}} 
		& \picturewithcoordinates{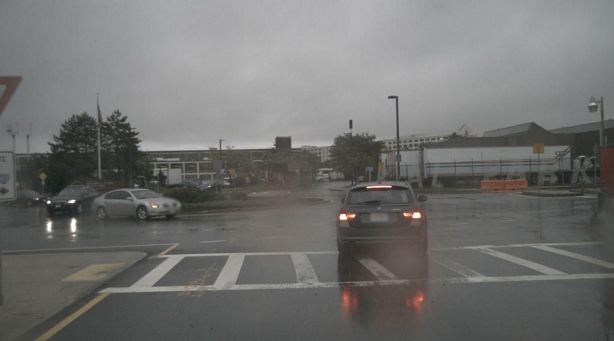}{(0, -1.35)}{(0, -0.85)}{(0, -1.35)}{(-1.0, -1.35)}
		& \picturewithcoordinates{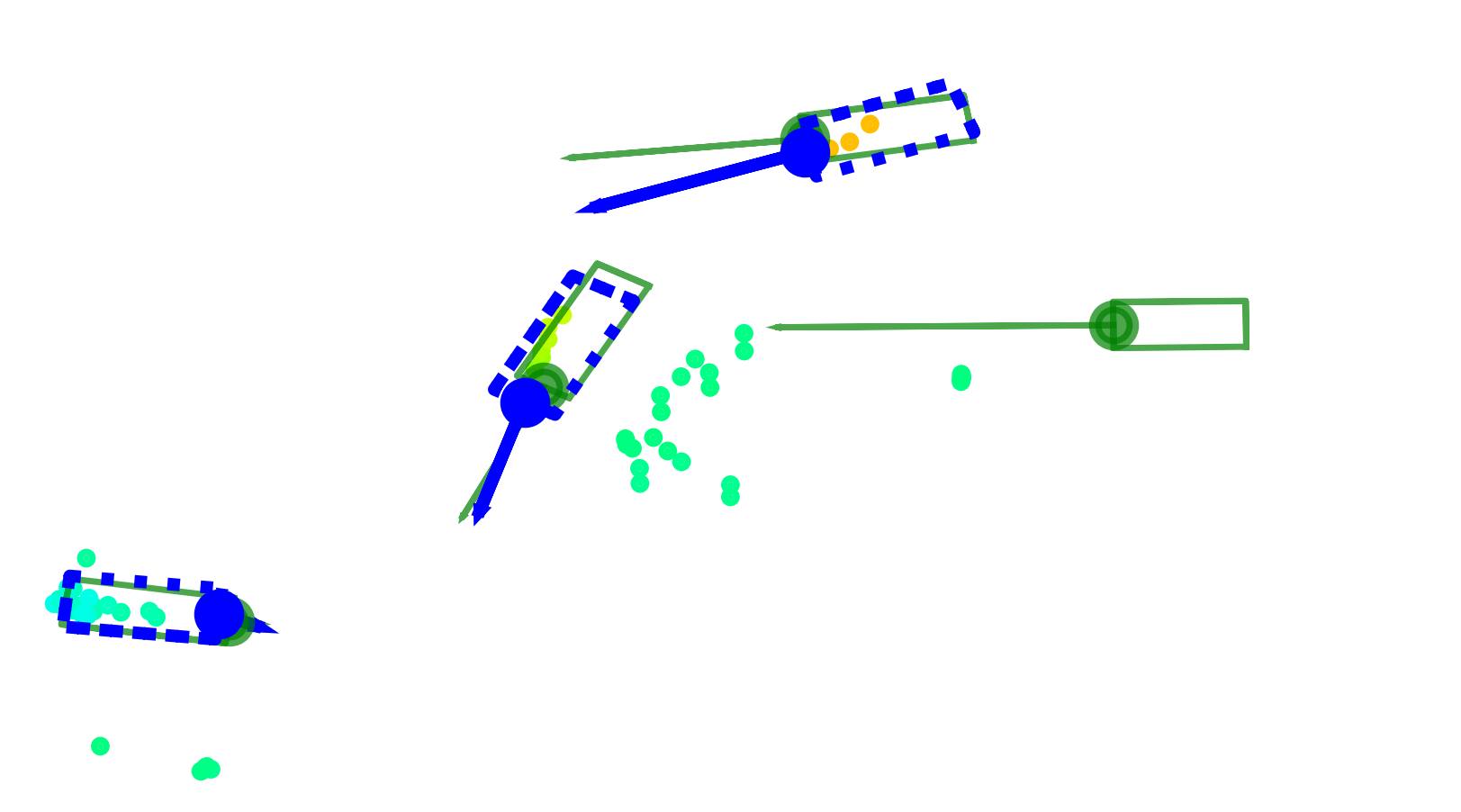}{(-2.5, -0.4)}{(-2.0, -0.4)}{(-2.5, -0.4)}{(-2.5, 0.6)}
		& \picturewithcoordinates{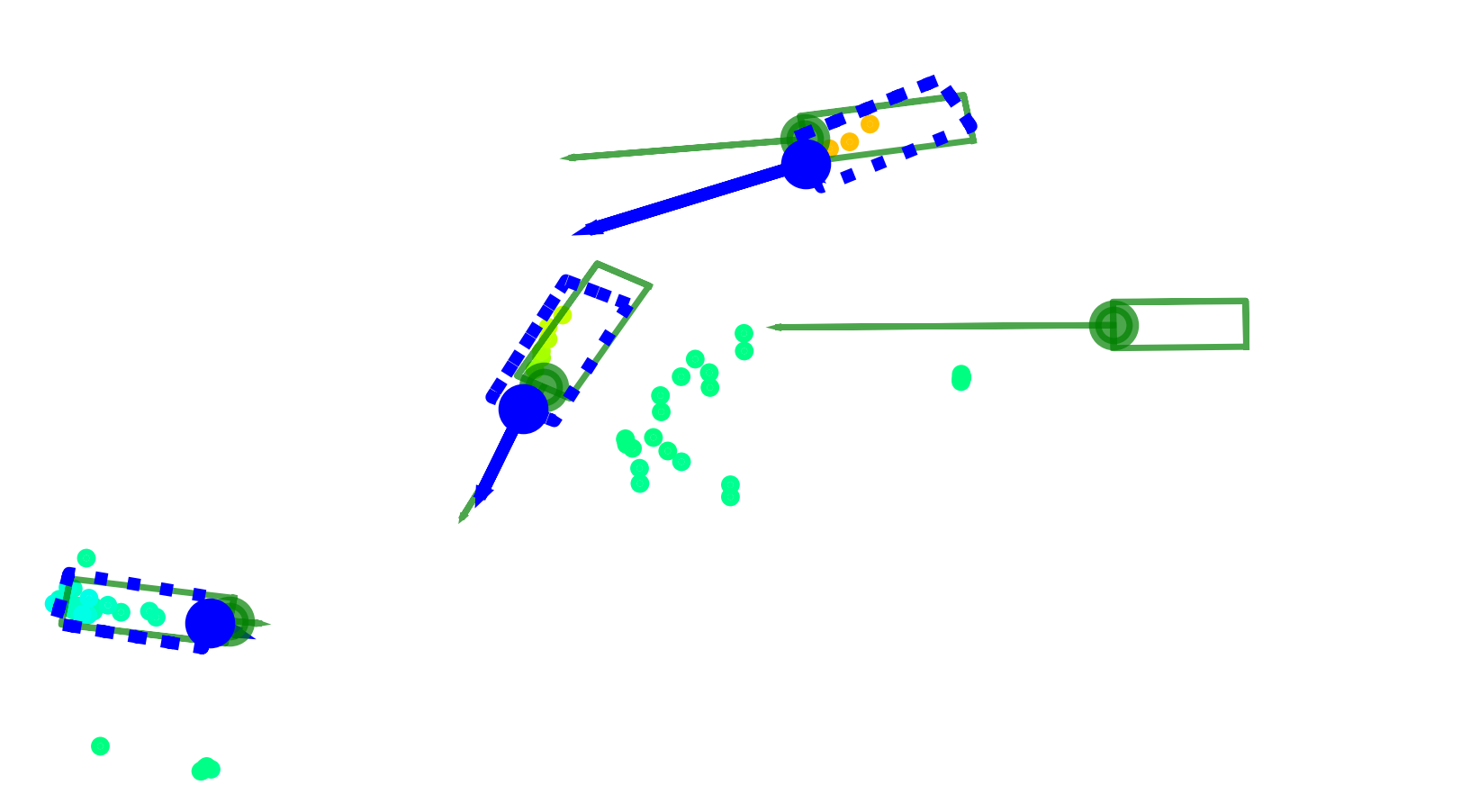}{(-2.5, -0.4)}{(-2.0, -0.4)}{(-2.5, -0.4)}{(-2.5, 0.6)}
		\\ \hline
		\raisebox{0.85\totalheight}{\rotatebox[origin=c]{90}{Example $3$}}
		& \picturewithcoordinates{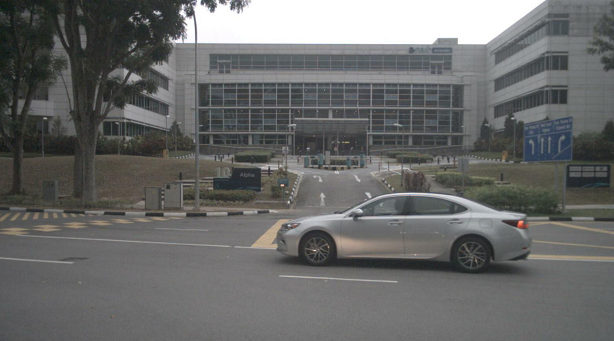}{(0, -1.35)}{(0, -0.85)}{(0, -1.35)}{(-1.0, -1.35)}
		& \picturewithcoordinates{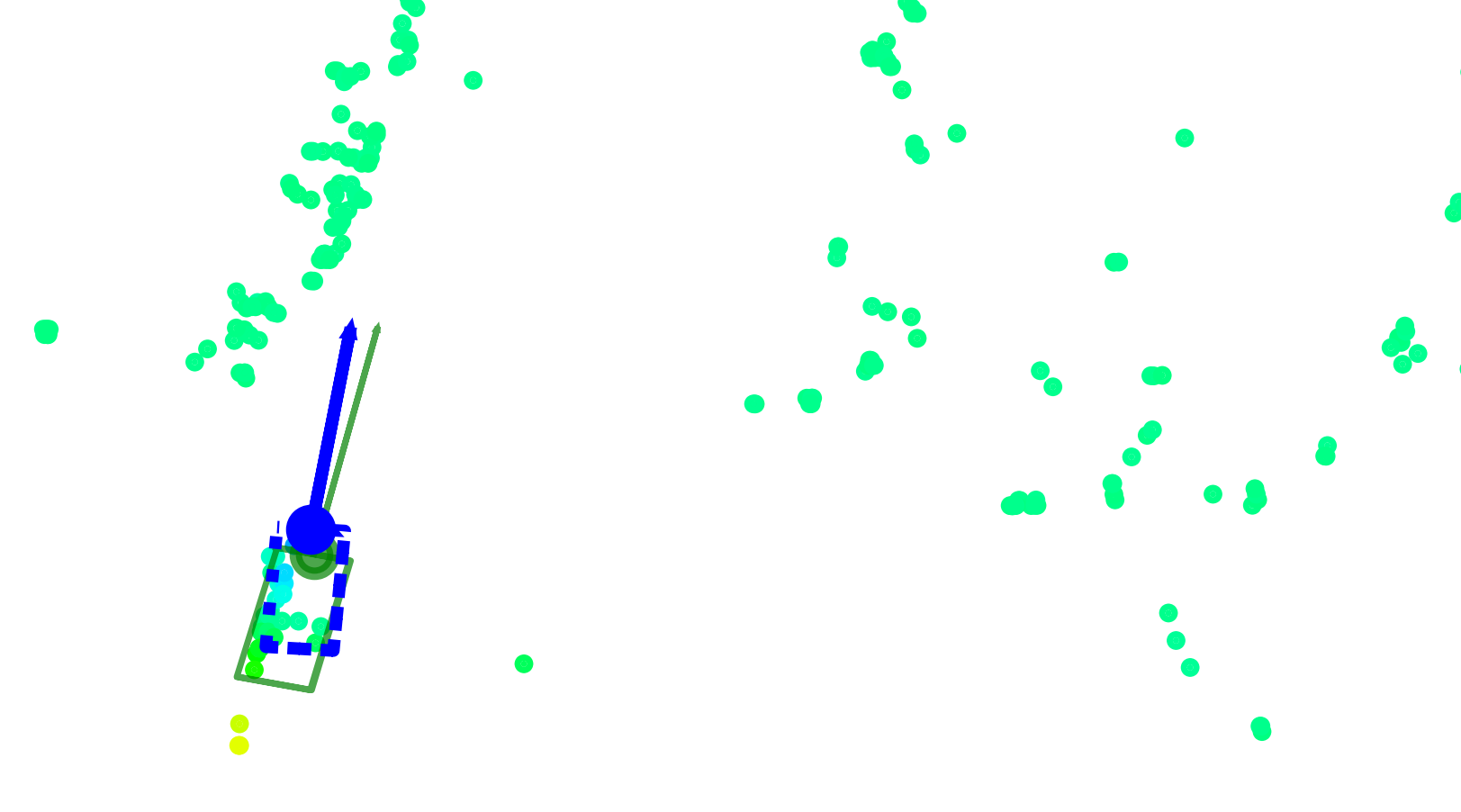}{(-2.5, -0.4)}{(-2.0, -0.4)}{(-2.5, -0.4)}{(-2.5, 0.6)}
		& \picturewithcoordinates{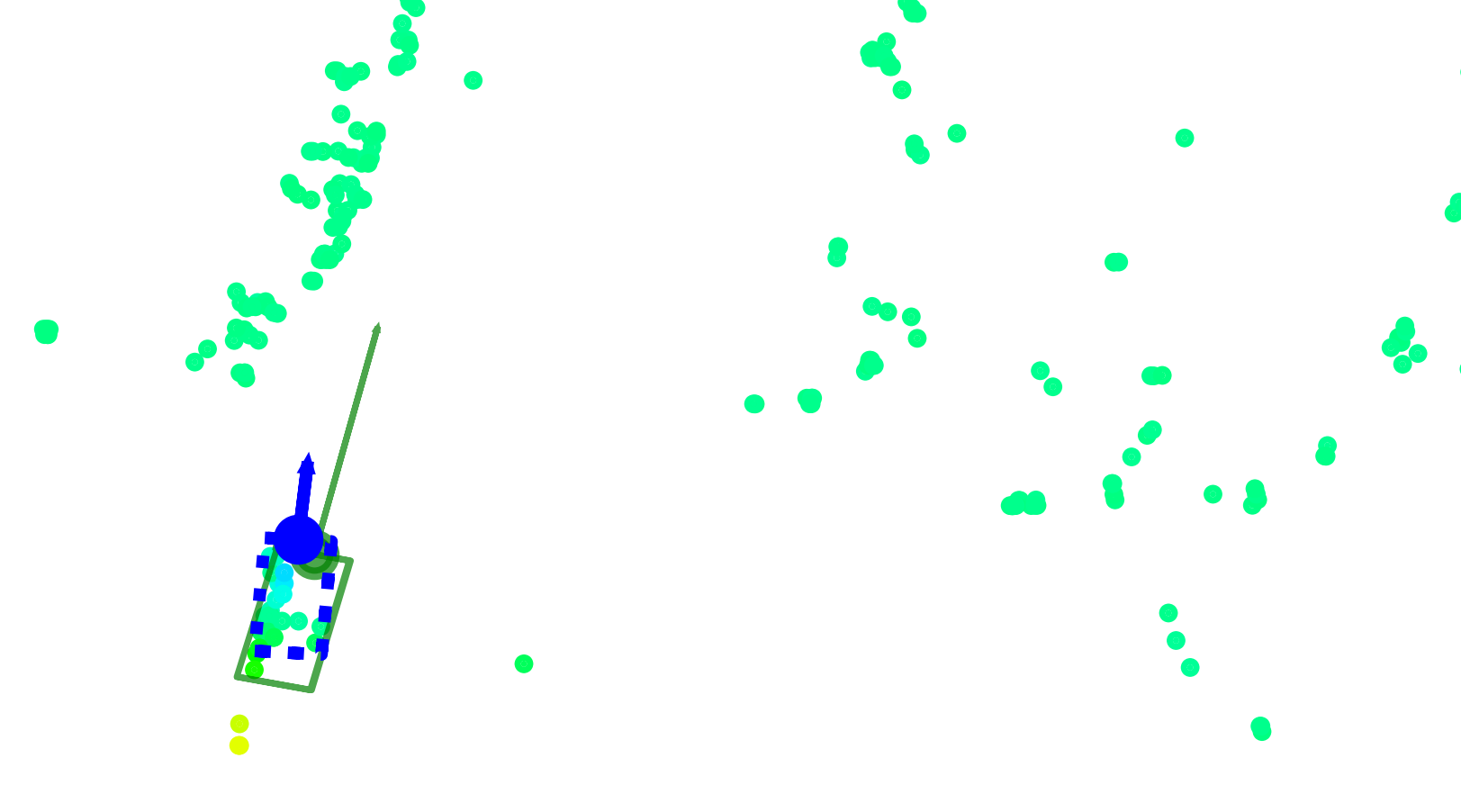}{(-2.5, -0.4)}{(-2.0, -0.4)}{(-2.5, -0.4)}{(-2.5, 0.6)}
		\\ \hline
		\raisebox{0.85\totalheight}{\rotatebox[origin=c]{90}{Example $4$}}
		& \picturewithcoordinates{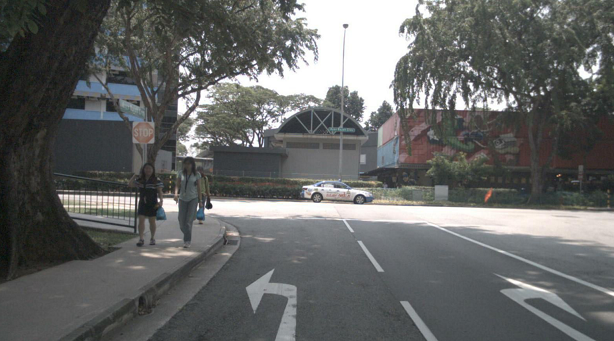}{(0, -1.35)}{(0, -0.85)}{(0, -1.35)}{(-1.0, -1.35)}
		& \picturewithcoordinates{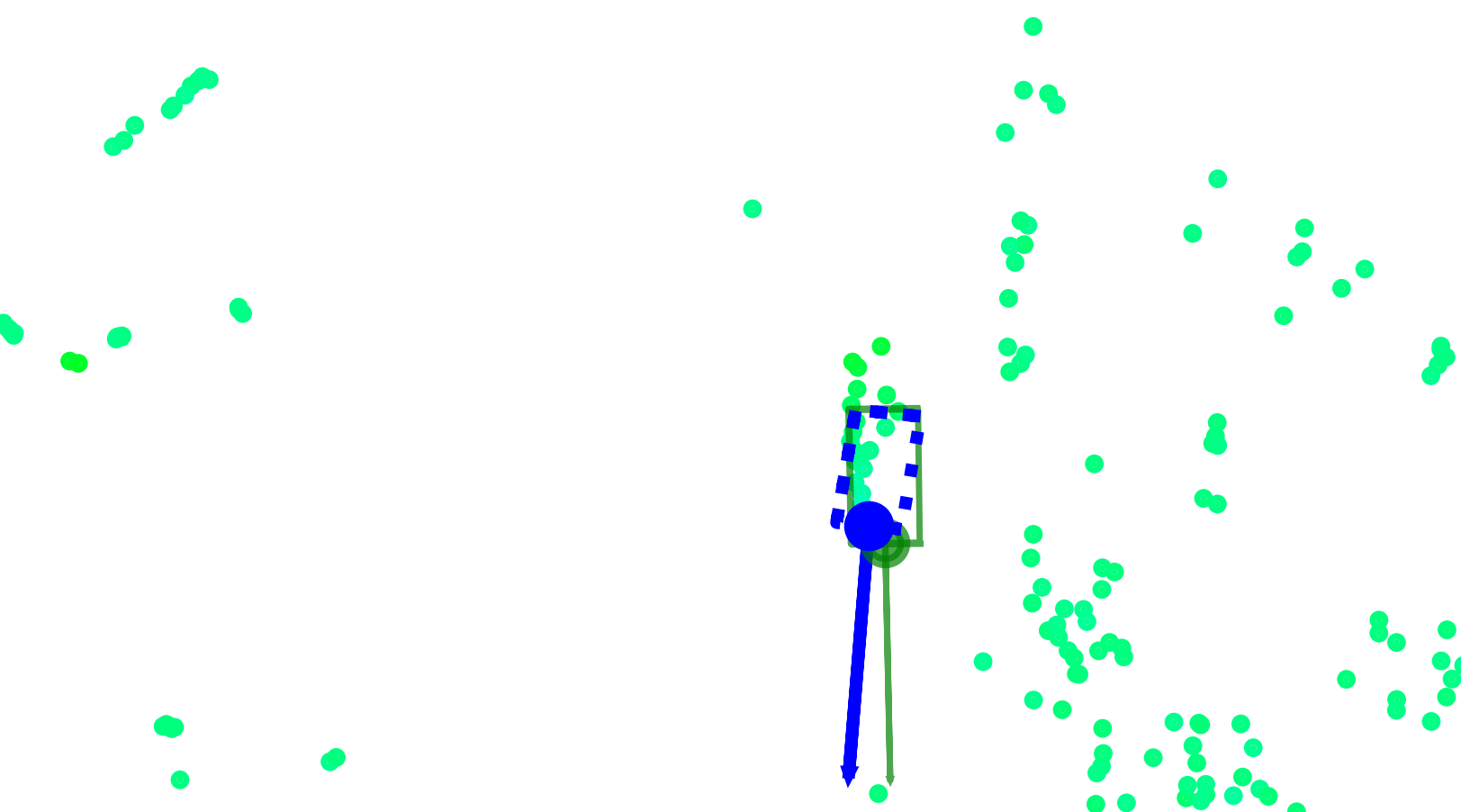}{(-2.5, -0.4)}{(-2.0, -0.4)}{(-2.5, -0.4)}{(-2.5, 0.6)}
		& \picturewithcoordinates{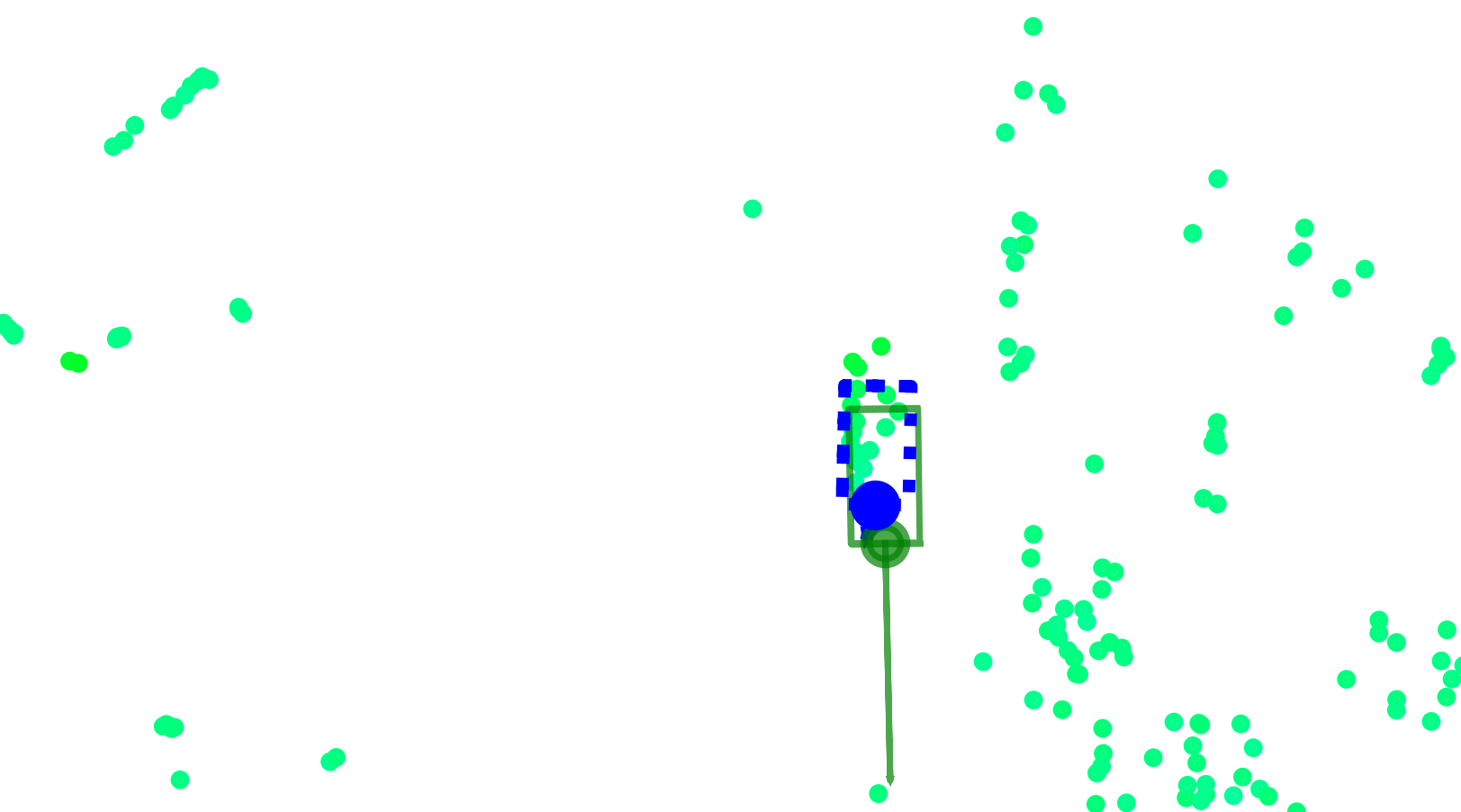}{(-2.5, -0.4)}{(-2.0, -0.4)}{(-2.5, -0.4)}{(-2.5, 0.6)}
		
	\end{tabular}
	
	\caption{Camera images and a bird eye view plot of the radar point clouds and network outputs of our self-supervised approach vs. the projected Doppler lower performance boundary in different nuScenes samples. Radar reflections are represented as colored dots. The color encodes the ego motion compensated Doppler velocity, where green corresponds to no radial motion. 
	Predicted boxes are drawn with dashed blue lines \protect\includegraphics[]{figures/predbox.pdf} and ground truth boxes with solid green lines \protect\includegraphics[]{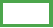}. 
	The velocities are drawn as oriented arrows with a length proportional to the absolute speed in the same color as the boxes. 
	It can be seen that our self-supervised training significantly outperforms the Doppler approach for tangential motions to the radar sensor (see velocity differences in Example $3$ and $4$), while there is almost no difference between the two approaches for radial movements (see longitudinal objects in Example $1$ and $2$). Furthermore, in Example $2$ there is a ground truth box that could not be detected due to a lack of radar measurements.}\label{fig:exemplary_results}
	\vspace{1em}
\end{figure*}
\begin{table*}[!t]
	\centering
	\caption{Benchmark of our network using different velocity targets evaluated on the nuScenes validation dataset for class \textit{car}
	}
	\label{tab:benchmark_nuscenes}
	\begin{tabular}{l c c c c c}
		\toprule
		\textbf{Velocity target} & \shortstack{\textbf{requires} \\ \textbf{OBB labels}} & \shortstack{\textbf{requires} \\ \textbf{velocity labels}} & \textbf{AP $(\%)$ $\uparrow$} & \textbf{AP$\mathbf{4.0}$ $(\%)$ $\uparrow$} & \textbf{AVE $(\si{m/s})$ $\downarrow$} \\
		\hline
		label & yes & yes &23.47 & 38.45 & \textbf{0.480}\\
		self-supervised (ours) & yes & \textbf{no} &23.49& 39.03 & 0.515\\
		projected Doppler & yes & \textbf{no}& \textbf{23.86} & \textbf{39.29} & 0.580\\
		\bottomrule
	\end{tabular}
\end{table*}

\begin{table}[!t]
	\centering
	\caption{Ablation study: impact of different network extensions evaluated on the nuScenes validation dataset for class \textit{car}
	}
	\label{tab:ablation_study_network}
	\begin{tabular}{l c c c}
		\toprule
		\textbf{Network configuration} & \textbf{AP $(\%)$ $\uparrow$} & \textbf{AP$\mathbf{4.0}$ $(\%)$ $\uparrow$} &  \textbf{AVE $(\si{m/s})$ $\downarrow$} \\
		\hline
		no $v_r$ pre-training & 22.31 & 38.36 & 0.797\\
		no TemporalPillars & 22.18 & 37.55 & 0.561\\
		no $v_r$-map & 23.23 & 38.71 & 0.541\\
		Proposed (ours) & \textbf{23.49}& \textbf{39.03} & \textbf{0.515}\\
		\bottomrule
	\end{tabular}
\end{table}

\begin{table}[!t]
	\centering
	\caption{Ablation study: different number of aggregated scans evaluated on the nuScenes validation dataset for class \textit{car}
	}
	\label{tab:ablation_frame_aggregation}
	\begin{tabular}{l c c c}
		\toprule
		\textbf{Number of scans} & \textbf{AP $(\%)$ $\uparrow$}  & \textbf{AP$\mathbf{4.0}$ $(\%)$ $\uparrow$} & \textbf{AVE $(\si{m/s})$ $\downarrow$} \\
		\hline
		1 & 14.67& 27.93 & 0.727\\
		3 & 19.77 & 34.67 & 0.638\\
		5 & 22.25 & 37.53 & 0.600\\
		7 & \textbf{23.49}& \textbf{39.03} & \textbf{0.515}\\
		\bottomrule
	\end{tabular}
\end{table}
The results are summarized in Tab.\,\ref{tab:benchmark_nuscenes}. 
It can be observed, that the proposed method achieves a lower AVE than the training with projected Doppler velocities and almost reaches the performance bound of the supervised velocity vector training. 
The AVE averages over all true positives, where many samples are simple stationary objects or longitudinally moving traffic. 
The Cartesian velocities in such cases are already well predicted by the projected Doppler approach. 
Hence, an AVE improvement of $0.065$ \si{m/s} is quite significant, considering that a large gain is only achievable for a small number of OBBs.

Examples are shown in Fig.\,\ref{fig:exemplary_results}, where the approach using the projected Doppler fails predicting the velocities of tangentially moving objects, while the presented self-supervised approach can handle the scenarios, even in difficult situations such as roundabouts or at intersections.
However, our model sometimes tends to predict an inaccurate velocity estimate (direction and absolute value) when only a few radar points are measured on an object, cf. Fig.~\ref{fig:exemplary_results} (Example $1$ and $2$).
This is probably due to the low information density and the high position variance typical for radar measurements.
The detection performance of the three methods is similar, as expected. 
The AP and AP$4.0$ is slightly better for the projected Doppler-based training, where the proposed methods marginally outperforms the fully supervised training.
However, the difference in detection performance is not significant.

In addition, we conducted ablation studies to investigate the impact of the network extensions described in Sec.\,\ref{sec:velocity_extension} and Sec.\,\ref{sec:training_phases} as well as the impact of the number of scans aggregated in the input point cloud. 
The results for different configurations are shown in Tab.\,\ref{tab:ablation_study_network}, where 
\textit{Proposed} refers to the self-supervised approach as described in this paper.
In this experiment, single components are omitted, i.e. the Doppler-based pre-training, the TemporalPillars and the $v_r$-map. 
The results show that the Doppler-based pre-training has a high impact on the velocity estimation. 
Clearly, the network architecture is the same with and without this pre-training, but training convergence is highly affected. 
We have observed that $\mathcal{L}_\text{vel}$ can be quite noisy, which makes a good initialization of the network weights necessary. 
Next, the use of TemporalPillars reduces the velocity error by roughly $8\%$, in comparison to merging all scans in a single point cloud. 
Further, we observe that the TemporalPillars also improves the detection performance. 
Similarly, the proposed $v_r$-map as input to the backbone and through the shortcut  improved the velocity estimation. 

A second ablation study investigates the impact of the number of radar scans which are aggregated in one frame in Tab.\,\ref{tab:ablation_frame_aggregation}. 
We aggregated $1$, $3$, $5$ and $7$ scans, corresponding to a time interval of \SI{0}{ms}, \SI{150}{ms}, \SI{300}{ms} and \SI{450}{ms}, respectively, for the nuScenes radar sensors.
Obviously, a larger aggregation interval improves both detection performance and velocity estimates, but also introduces more delay and temporal correlation in a subsequent tracking system. 
In addition, the shape and position information of objects can be distorted, especially if they are moving.
Interestingly, the reduction from $7$ to $5$ scans increases the velocity error significantly. 
We believe that this might be because the radar point reflections on the vehicles are not at consistent places but at random locations on the car. 
A short history (trace) might not be sufficient to observe the overall trend in which direction the object is moving. 
Furthermore, it can be observed that velocity estimation is also possible with a single scan. 
Clearly, the history (trace) of reflection points improves the performance, but usable information is also present in the radial velocity measurements. 

\section{Conclusion}
In this paper, we presented a self-supervised velocity regression approach for single-shot radar object detection networks and proposed a suitable network architecture.
The training consists of two parts: 
A supervised box detection part and a self-supervised velocity regression part. 
A combination of both parts results in a network capable of predicting oriented bounding boxes and corresponding Cartesian velocities without the need for sequence or velocity labels.
Our experiments on the publicly available nuScenes dataset showed that our method can achieve comparable results to a fully supervised training. 
The proposed approach is particularly useful when velocity labels are not available.
Labels for Cartesian velocity would be expensive to obtain, but are not required by the proposed method.  
Possible extensions of this work could use the proposed self-supervised velocity estimation for pre-training or as an auxiliary task. 
In addition, motion models other than the constant velocity model could be tested and the time interval between \textit{detection step} and \textit{velocity step} $\Delta t_{\text{vel} \rightarrow \text{det}}$ could be varied during training, which might increase the robustness of the velocity estimate.

\addtolength{\textheight}{-7.1cm}   


\bibliographystyle{ieeetr}
\bibliography{references}

\end{document}

%% file: figures/teaser.tex
\newcommand{\filmhole}{
	\begin{tikzpicture}
	\node at (0,0) [rounded corners=.5mm, fill=white, minimum width=1.8mm, minimum height=2.8mm, outer sep=1pt, inner sep=0pt] {};
	\end{tikzpicture}}

\newcommand{\filmsample}[1]{
	\begin{tikzpicture}
	\matrix (film) [fill=black, draw=black, matrix of nodes, nodes in empty cells, ampersand replacement=\&,column sep=.3em, row sep=.7em, outer sep=0pt,inner sep=.3pt] {%
		{} \& \filmhole \& \filmhole \& \filmhole \& \filmhole \& \filmhole \& \filmhole \& \filmhole \& \filmhole \\
		{} \& {} \& {} \& {} \& {} \& {} \& {} \& {} \& {}\\
		{} \& {} \& {} \& {} \& {} \& {} \& {} \& {} \& {}\\
		{} \& {} \& {} \& {} \& {} \& {} \& {} \& {} \& {}\\
		{} \& {} \& {} \& {} \& {} \& {} \& {} \& {} \& {}\\
		{} \& {} \& {} \& {} \& {} \& {} \& {} \& {} \& {}\\
		{} \& \filmhole \& \filmhole \& \filmhole \& \filmhole \& \filmhole \& \filmhole \& \filmhole \& \filmhole\\
	};
	\node[rounded corners=4pt, fill=white, fit={([shift={(2pt,5pt)}]film-2-1.north west)([shift={(-2pt,-5pt)}]film-6-9.south east)}, path picture={
		\node at (path picture bounding box.south){
			\includegraphics[width=1.3\columnwidth]{#1}
		};}] {};
	\end{tikzpicture}}

\begin{tikzpicture}[scale=.9]

\node (film1) [inner sep=0pt] {\filmsample{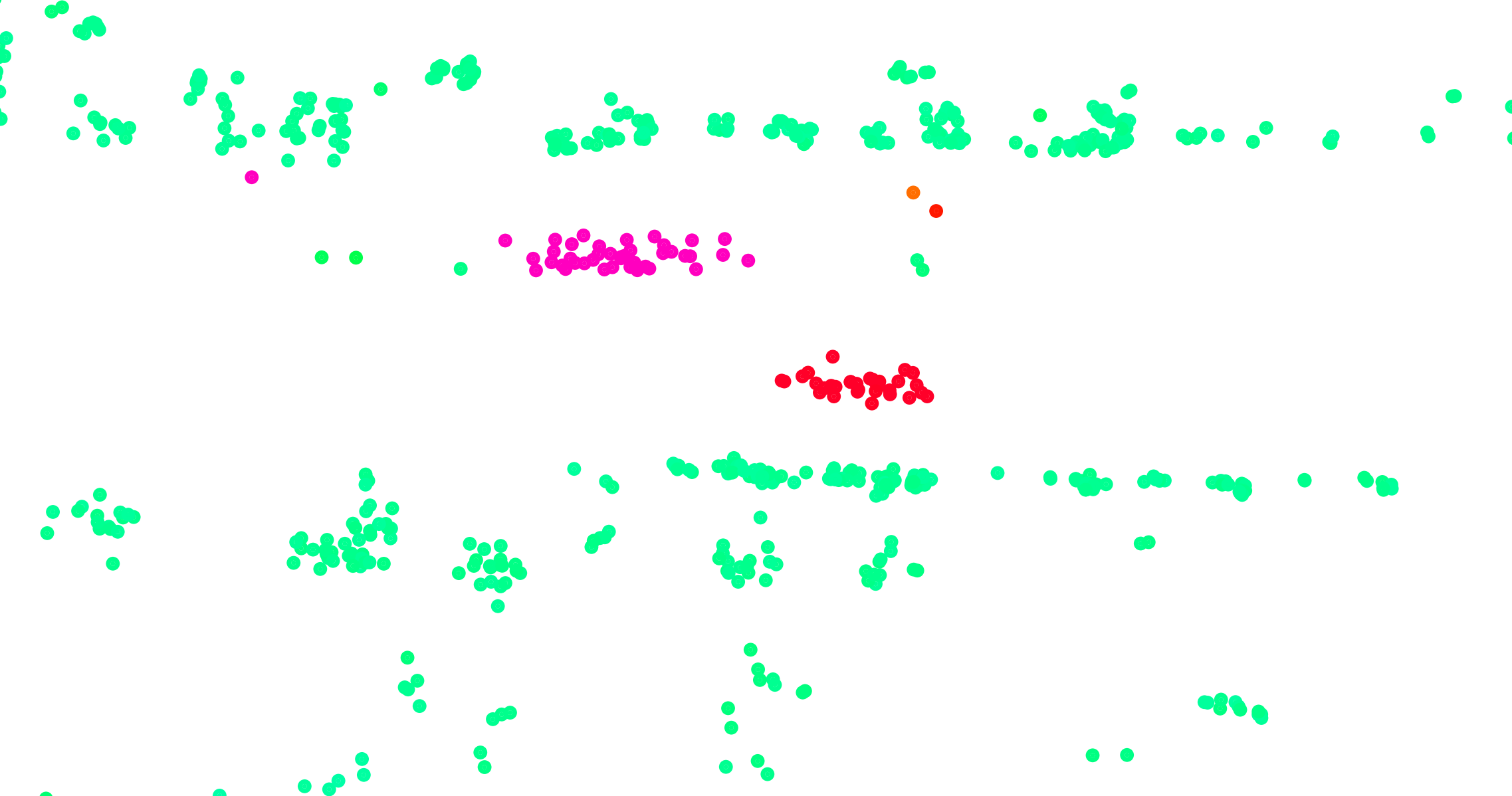}};
\node (film2) at (film1.east) [anchor=west, inner sep=0pt] {\filmsample{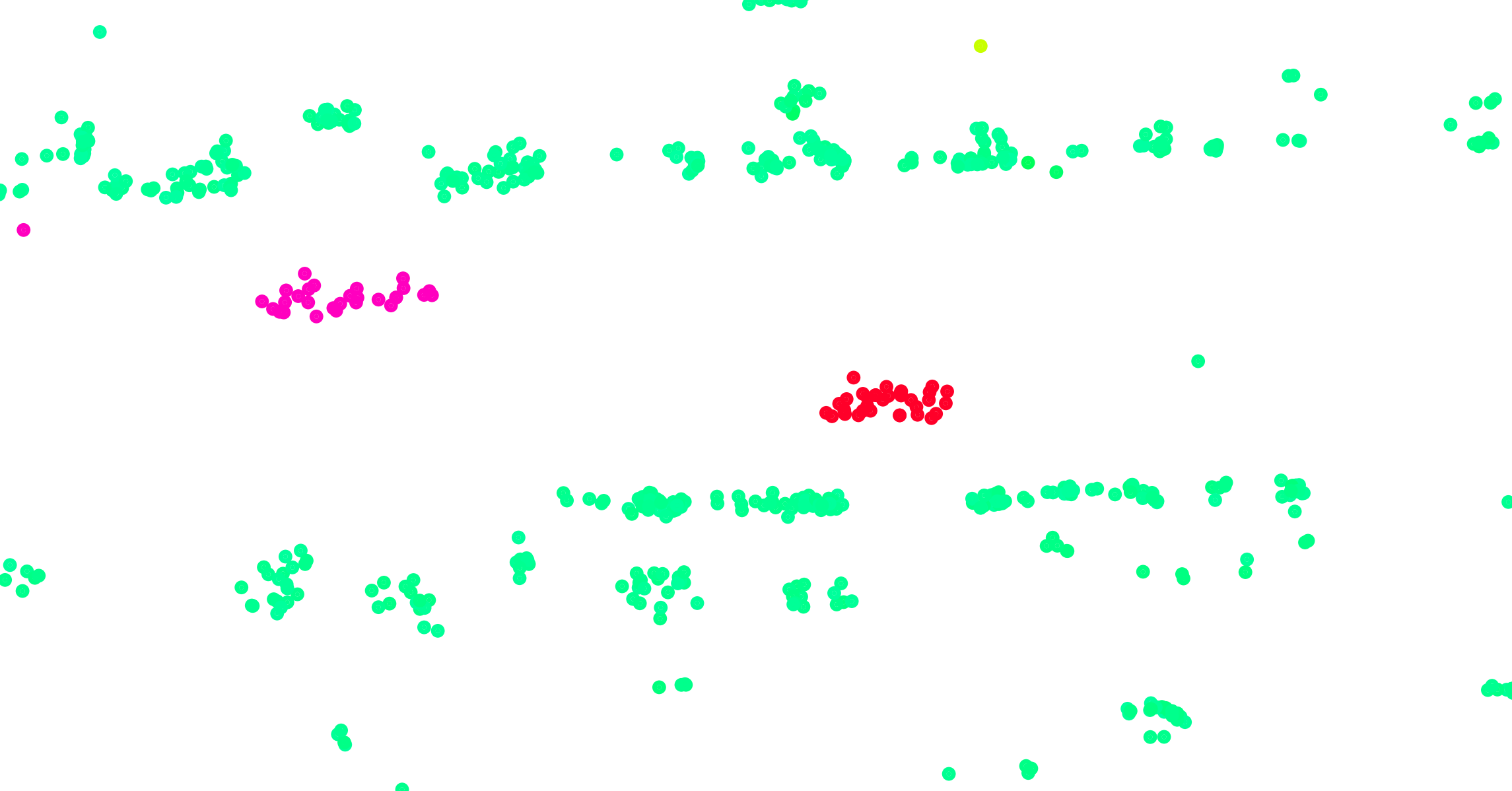}};
\node (film3) at (film2.east) [anchor=west, inner sep=0pt] {\filmsample{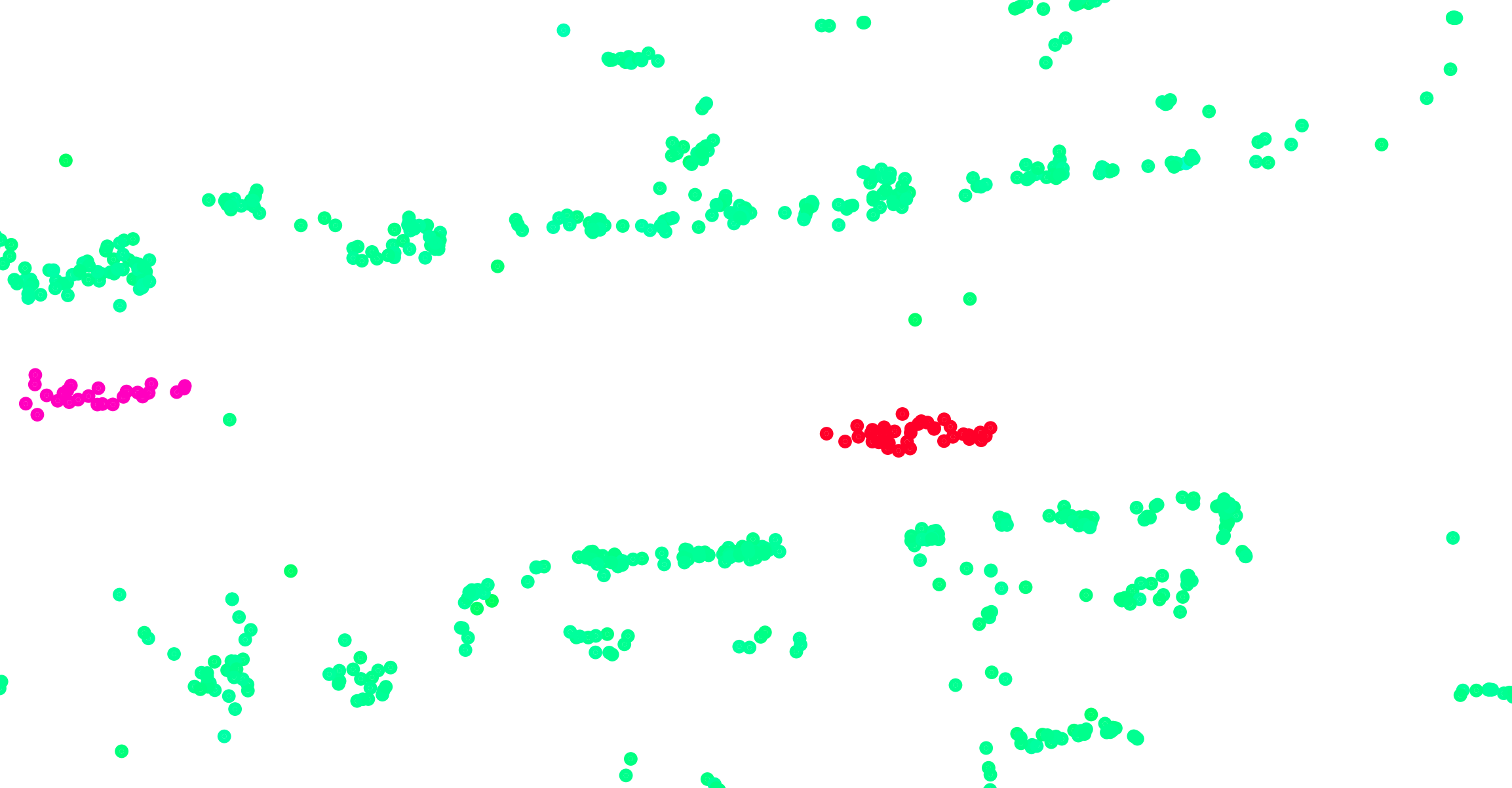}};

\node (pointcloud_label) at ($(film1.west) + (-.2,0)$) [rotate=90] {\footnotesize point cloud};

\draw[thick, -{stealth}] ($(film1.north west) + (0,0.2)$) -- ($(film3.north east) + (-.5,0.2)$) node [right] {$t$};

\node (detnet_t0) [below=0.2 of film1] {
	\begin{tikzpicture}[scale=1.1]
	\draw (-1.1,.8) -- (1.1,.8) -- (1.1,-.7) -- (0,-1.0) -- (-1.1,-.7) -- cycle;
	\begin{scope}[scale=.5, shift={(0,.5)}, every node/.append style={
		yslant=0.5,xslant=-1},yslant=0.5,xslant=-1
	]

	\draw[fill=white] ($(-.85,-.85)+(-1.5,-1.5)$) rectangle ($(.85,.85)+(-1.5,-1.5)$);
	\draw[black,very thick] ($(-.85,-.85)+(-1.5,-1.5)$) rectangle ($(.85,.85)+(-1.5,-1.5)$);
	
	\draw[fill=white] ($(-.7,-.7)+(-1.2,-1.2)$) rectangle ($(.7,.7)+(-1.2,-1.2)$);
	\draw[black,very thick] ($(-.7,-.7)+(-1.2,-1.2)$) rectangle ($(.7,.7)+(-1.2,-1.2)$);

	\draw[fill=white] ($(-.7,-.7)+(-0.9,-0.9)$) rectangle ($(.7,.7)+(-0.9,-0.9)$);
	\draw[black,very thick] ($(-.7,-.7)+(-0.9,-0.9)$) rectangle ($(.7,.7)+(-0.9,-0.9)$);
	
	\draw[fill=white] ($(-.85,-.85)+(-0.6,-0.6)$) rectangle ($(.85,.85)+(-0.6,-0.6)$);
	\draw[black,very thick] ($(-.85,-.85)+(-0.6,-0.6)$) rectangle ($(.85,.85)+(-0.6,-0.6)$);
	
	\draw[fill=white] ($(-1,-1)+(-0.3,-0.3)$) rectangle ($(1,1)+(-0.3,-0.3)$);
	\draw[black,very thick] ($(-1,-1)+(-0.3,-0.3)$) rectangle ($(1,1)+(-0.3,-0.3)$);

	\fill[white,fill opacity=0.9] (-1,-1) rectangle (1,1);
	
	\fill[red] (0.6,0.6) rectangle (0.4,0.4); 
	\fill[blue] (-0.2,0.2) rectangle (-0.4,0.0); 
	\fill[orange] (-0.4,-0.6) rectangle (-0.6,-0.8); 
	\fill[red!50!white] (-0.2,0.0) rectangle (0.0,0.2); 
	\fill[blue!50!white] (0.2,-0.8) rectangle (0.0,-1.0); 
	\fill[orange!50!white] (0.0,-0.2) rectangle (-0.2,0.0);  
	
	\draw[step=.2, black] (-1,-1) grid (1,1); 
	\draw[black,very thick] (-1,-1) rectangle (1,1);
	\end{scope}
	\end{tikzpicture}	
};
\draw [draw=none] (pointcloud_label) |- node [midway, rotate=90] {\footnotesize detection network}(detnet_t0.west);

\node (obb_t0) [below right = -.2 and -1.2 of detnet_t0.south] {
	\begin{tikzpicture}[scale=.8,yslant=0.5,xslant=-1]
	\draw [draw] (-.7,1) -- (.9,1) -- (.9,-1.1) -- (-.7,-1.1) -- cycle;
	\draw [blue, densely dotted, line width = 1pt] (-.3,.3) rectangle (.4,.6);
	\draw [blue, line width = 1pt, -{stealth}] (.4,.45) --(0.9,.45);
	\draw [blue, densely dotted, line width = 1pt, rotate=-50, anchor=center] (.8,-.3) rectangle (.1,-.6);
	\end{tikzpicture}	
};
\node [below=-.2 of obb_t0.south, anchor=north, align=center] {\footnotesize OBBs at $t_0-\Delta t$};

\node (detnet_t1) [below = 0.2 of film3] {
	\begin{tikzpicture}[scale=1.1]
	\draw (-1.1,.8) -- (1.1,.8) -- (1.1,-.7) -- (0,-1.0) -- (-1.1,-.7) -- cycle;
	\begin{scope}[scale=.5, shift={(0,.5)}, every node/.append style={
		yslant=0.5,xslant=-1},yslant=0.5,xslant=-1
	]

	\draw[fill=white] ($(-.85,-.85)+(-1.5,-1.5)$) rectangle ($(.85,.85)+(-1.5,-1.5)$);
	\draw[black,very thick] ($(-.85,-.85)+(-1.5,-1.5)$) rectangle ($(.85,.85)+(-1.5,-1.5)$);
	
	\draw[fill=white] ($(-.7,-.7)+(-1.2,-1.2)$) rectangle ($(.7,.7)+(-1.2,-1.2)$);
	\draw[black,very thick] ($(-.7,-.7)+(-1.2,-1.2)$) rectangle ($(.7,.7)+(-1.2,-1.2)$);

	\draw[fill=white] ($(-.7,-.7)+(-0.9,-0.9)$) rectangle ($(.7,.7)+(-0.9,-0.9)$);
	\draw[black,very thick] ($(-.7,-.7)+(-0.9,-0.9)$) rectangle ($(.7,.7)+(-0.9,-0.9)$);
	
	\draw[fill=white] ($(-.85,-.85)+(-0.6,-0.6)$) rectangle ($(.85,.85)+(-0.6,-0.6)$);
	\draw[black,very thick] ($(-.85,-.85)+(-0.6,-0.6)$) rectangle ($(.85,.85)+(-0.6,-0.6)$);
	
	\draw[fill=white] ($(-1,-1)+(-0.3,-0.3)$) rectangle ($(1,1)+(-0.3,-0.3)$);
	\draw[black,very thick] ($(-1,-1)+(-0.3,-0.3)$) rectangle ($(1,1)+(-0.3,-0.3)$);

	\fill[white,fill opacity=0.9] (-1,-1) rectangle (1,1);
	
	\fill[red] (0.6,0.6) rectangle (0.4,0.4); 
	\fill[blue] (0.0,0.0) rectangle (-0.2,-0.2); 
	\fill[orange] (-0.4,-0.6) rectangle (-0.6,-0.8); 
	\fill[red!50!white] (0.0,-0.2) rectangle (+0.2,0.0); 
	\fill[blue!50!white] (0.2,-0.8) rectangle (0.0,-1.0); 
	\fill[orange!50!white] (0.2,-0.4) rectangle (0.0,-0.2);  
	
	\draw[step=.2, black] (-1,-1) grid (1,1); 
	\draw[black,very thick] (-1,-1) rectangle (1,1);
	\end{scope}
	\end{tikzpicture}	
};
\node (obb_t1) [below left = -.2 and -1.2 of detnet_t1.south] {
	\begin{tikzpicture}[scale=.8,yslant=0.5,xslant=-1]
	\draw [draw] (-.7,1) -- (.9,1) -- (.9,-1.1) -- (-.7,-1.1) -- cycle;
	\draw [blue, densely dotted, line width = 1pt] (0.2,.3) rectangle (.9,.6);
	\draw [blue!30!green, densely dotted, line width = 1pt] (0.1,.2) rectangle (.8,.5);
	
	\draw [blue, densely dotted, line width = 1pt, rotate=-50, anchor=center] (.8,-.3) rectangle (.1,-.6);
	\draw [blue!30!green, densely dotted, line width = 1pt, rotate=-50, anchor=center] (.75,-.4) rectangle (.05,-.7);
	\end{tikzpicture}	
};
\node [below=-.2 of obb_t1.south, anchor=north, align=center] {\footnotesize OBBs at $t_0$};
\draw [draw=none] (pointcloud_label) |- node [midway, rotate=90, align=center, font=\footnotesize] {predicted \\ OBBs}(obb_t0.west);

\draw [-{stealth}, line width=1pt] (obb_t0.east) to [bend left] node [midway, above] {$+\Delta t$} (obb_t1.west);

\end{tikzpicture}

%% file: figures/network_architecture_vel.tex
\begin{tikzpicture}[scale=.9,every node/.style={minimum size=1},on grid]

\definecolor{point_color}{RGB}{0,255,132}

\node (pointcloud0) {
	\begin{tikzpicture}[scale=.4]
	\draw [draw=none] (-.7,1) -- (.7,1) -- (.7,-1) -- (-.7,-1) -- cycle;
	\fill[point_color] (.6,.7) circle (.35em);
	\fill[point_color] (0,0) circle (.35em);
	\fill[point_color] (-.1,.6) circle (.35em);
	\fill[point_color] (-.5,-.4) circle (.35em);
	\fill[point_color] (0,-.4) circle (.35em);
	\fill[point_color] (.7,-.2) circle (.35em);
	\end{tikzpicture}	
};
\node [left = 0.5 of pointcloud0] {\footnotesize $t_0$};
\node (pointcloud1) [below= -.2 of pointcloud0.south, anchor=north] {
	\begin{tikzpicture}[scale=.4]
	\draw [draw=none] (-.7,1) -- (.7,1) -- (.7,-1) -- (-.7,-1) -- cycle;
	\fill[point_color] (.5,.8) circle (.35em);
	\fill[point_color] (0,-.2) circle (.35em);
	\fill[point_color] (-.1,.6) circle (.35em);
	\fill[point_color] (-.6,-.3) circle (.35em);
	\fill[point_color] (0,-.4) circle (.35em);
	\fill[point_color] (.7,-.2) circle (.35em);
	\end{tikzpicture}	
};
\node [left = 0.5 of pointcloud1] {\footnotesize $t_1$};
\node (pcl_dots) [below=-.1 of pointcloud1.south] {...};
\node (pointcloud2) [below= 0 of pcl_dots, anchor=north] {
	\begin{tikzpicture}[scale=.4]
	\draw [draw=none] (-.7,1) -- (.7,1) -- (.7,-1) -- (-.7,-1) -- cycle;
	\fill[point_color] (.4,.3) circle (.35em);
	\fill[point_color] (-.2,-.2) circle (.35em);
	\fill[point_color] (-.2,.5) circle (.35em);
	\fill[point_color] (-.6,-.3) circle (.35em);
	\fill[point_color] (0,-.4) circle (.35em);
	\fill[point_color] (.7,-.2) circle (.35em);
	\end{tikzpicture}	
};
\node [left = 0.5 of pointcloud2] {\footnotesize $t_n$};

\node (temporalpillar) [above right = 0.65 and .55 of pointcloud0.east, anchor = north west, draw, rounded corners, dotted, minimum width=.22\textwidth, text width=.2\textwidth, text depth=.143\textwidth, align=left] {\footnotesize \textsl{TemporalPillars} (a)};

\node (render0) [right = .8 of pointcloud0.east, draw, rounded corners] {\footnotesize PP-render};
\node (render1) [right = .8 of pointcloud1.east, draw, rounded corners] {\footnotesize PP-render};
\node (render2) [right = .8 of pointcloud2.east, draw, rounded corners] {\footnotesize PP-render};
\node (render_points) at ($(render2.north)!0.5!(render1.south)$) {...};

\node (grid0) [right=0.4 of render0.east, anchor=west] {
	\begin{tikzpicture}[scale=1.1]
	\begin{scope}[scale=.5, every node/.append style={
		yslant=0.5,xslant=-1},yslant=0.5,xslant=-1
	]
	\fill[white,fill opacity=0.9] (-.6,-.6) rectangle (.6,.6);
	
	\fill[red] (0.4,0.4) rectangle (0.2,0.2); 
	\fill[blue] (-0.2,0.2) rectangle (-0.4,0.0); 
	\fill[orange] (-0.4,-0.4) rectangle (-0.6,-0.6); 
	\fill[red!50!white] (-0.4,0.0) rectangle (-0.2,0.2); 
	\fill[blue!50!white] (0.2,-0.4) rectangle (0.0,-.6); 
	\fill[orange!50!white] (0.0,-0.2) rectangle (-0.2,0.0);  
	
	\draw[step=.2, black] (-.6,-.6) grid (.6,.6); 
	\draw[black,very thick] (-.6,-.6) rectangle (.6,.6);
	\end{scope}
	\end{tikzpicture}	
};
\node (grid1) [right=0.4 of render1.east, anchor=west] {
	\begin{tikzpicture}[scale=1.1]
	\begin{scope}[scale=.5, every node/.append style={
		yslant=0.5,xslant=-1},yslant=0.5,xslant=-1
	]
	\fill[white,fill opacity=0.9] (-.6,-.6) rectangle (.6,.6);
	
	\fill[red] (0.6,0.6) rectangle (0.4,0.4); 
	\fill[blue] (-0.2,0.2) rectangle (-0.4,0.0); 
	\fill[orange] (-0.4,-0.4) rectangle (-0.6,-0.6); 
	\fill[red!50!white] (-0.2,0.0) rectangle (0.0,0.2); 
	\fill[blue!50!white] (0.2,-0.4) rectangle (0.0,-.6); 
	\fill[orange!50!white] (0.0,-0.2) rectangle (-0.2,0.0);  
	
	\draw[step=.2, black] (-.6,-.6) grid (.6,.6); 
	\draw[black,very thick] (-.6,-.6) rectangle (.6,.6);
	\end{scope}
	\end{tikzpicture}	
};
\node (grid2) [right=0.4 of render2.east, anchor=west] {
	\begin{tikzpicture}[scale=1.1]
	\begin{scope}[scale=.5, every node/.append style={
		yslant=0.5,xslant=-1},yslant=0.5,xslant=-1
	]
	\fill[white,fill opacity=0.9] (-.6,-.6) rectangle (.6,.6);
	
	\fill[red] (0.6,0.6) rectangle (0.4,0.4); 
	\fill[blue] (-0.2,0.2) rectangle (-0.4,0.0); 
	\fill[orange] (-0.4,-0.4) rectangle (-0.6,-0.6); 
	\fill[red!50!white] (-0.0,0.0) rectangle (0.2,0.2); 
	\fill[blue!50!white] (0.4,-0.4) rectangle (0.2,-.6); 
	\fill[orange!50!white] (0.0,-0.2) rectangle (-0.2,0.0);  
	
	\draw[step=.2, black] (-.6,-.6) grid (.6,.6); 
	\draw[black,very thick] (-.6,-.6) rectangle (.6,.6);
	\end{scope}
	\end{tikzpicture}	
};
\node (grid_points) at ($(grid2.north)!0.5!(grid1.south)$) {...};

\draw [solid,-{stealth}] (pointcloud0) -- (render0);
\draw [solid,-{stealth}] (pointcloud1) -- (render1);
\draw [solid,-{stealth}] (pointcloud2) -- (render2);
\draw [solid,-{stealth}] (render0) -- (grid0);
\draw [solid,-{stealth}] (render1) -- (grid1);
\draw [solid,-{stealth}] (render2) -- (grid2);

\draw [decorate,decoration={brace,amplitude=4pt}]
($(render2.south) - (1.1,.5)$) -- node [midway] (point_merger) {} ($(render2.south) - (1.1,1.1)$);

\node (render_vel) [right = .4 of point_merger.east, draw, rounded corners] {\footnotesize $v_r$-render};

\coordinate [right = .1 of render_vel.east] (vel_grid_split);

\node (velgrid) [below=-.2 of grid2.south, anchor=north] {
	\begin{tikzpicture}[scale=1.1]
	\begin{scope}[scale=.5, every node/.append style={
		yslant=0.5,xslant=-1},yslant=0.5,xslant=-1
	]
	\fill[white,fill opacity=0.9] (-.6,-.6) rectangle (.6,.6);
	\draw[black,very thick] (-.6,-.6) rectangle (.6,.6);
	\end{scope}
	\end{tikzpicture}	
};
\node at (velgrid.center) [anchor=center] {\footnotesize (b)};

\draw [rounded corners=3pt, solid,-{stealth}] ($(pointcloud2.east)+(.1,0)$) |- ($(point_merger)+(.0,-.15)$);
\draw [rounded corners=3pt, solid,-{stealth}] ($(pointcloud1.east)+(.2,0)$) |- ($(point_merger)+(.0,0)$);
\draw [rounded corners=3pt, solid,-{stealth}] ($(pointcloud0.east)+(.3,0)$) |- ($(point_merger)+(.0,.15)$);
\draw [solid,-{stealth}] ($(point_merger)+(.25,.0)$) -- (render_vel);
\draw [solid] (render_vel.east) -- (vel_grid_split);
\draw [rounded corners=3pt, solid,-{stealth}] (vel_grid_split) |- (velgrid);

\draw [decorate,decoration={brace,amplitude=4pt}]
($(grid0.east) + (0,.2)$) -- node [midway] (temp_vel_concat) {} ($(velgrid.east) + (0,-.2)$);

\node (stem) [below right = 1 and .7 of temp_vel_concat, draw, rounded corners] {\footnotesize stem};

\begin{scope}[local bounding box=resgrid1,shift={($(stem.east)+(2.2,0)$)}, every node/.append style={
		yslant=0,xslant=1},yslant=0,xslant=1
	]
	\fill[white,fill opacity=0.9] (-1,-.4) rectangle (1,.4);
	\draw[black,very thick] (-1,-.4) rectangle (1,.4);
\end{scope}

\node (res2) [above = .7 of resgrid1, draw, rounded corners] {\footnotesize ResNet block 2};
\node (res2_num) [circle,draw=black, fill=black, text=white, inner sep=0pt,minimum size=5pt] at ($(res2.west)-(0.2,0)$) {\footnotesize $\mathbf{3\times}$};

\begin{scope}[local bounding box=resgrid2,shift={($(res2.north)+(0,.5)$)}, every node/.append style={
		yslant=0,xslant=1},yslant=0,xslant=1
	]
	\fill[white,fill opacity=0.9] (-.8,-.32) rectangle (.8,.32);
	\draw[black,very thick] (-.8,-.32) rectangle (.8,.32);
\end{scope}

\node (res3) [above = .7 of resgrid2, draw, rounded corners] {\footnotesize ResNet block 3};
\node (res3_num) [circle,draw=black, fill=black, text=white, inner sep=0pt,minimum size=5pt] at ($(res3.west)-(0.2,0)$) {\footnotesize $\mathbf{6\times}$};

\begin{scope}[local bounding box=resgrid3,shift={($(res3.north)+(0,.4)$)}, every node/.append style={
		yslant=0,xslant=1},yslant=0,xslant=1
	]
	\fill[white,fill opacity=0.9] (-.6,-.24) rectangle (.6,.24);
	\draw[black,very thick] (-.6,-.24) rectangle (.6,.24);
\end{scope}

\node (res4) [above = .7 of resgrid3, draw, rounded corners] {\footnotesize ResNet block 4};
\node (res4_num) [circle,draw=black, fill=black, text=white, inner sep=0pt,minimum size=5pt] at ($(res4.west)-(0.2,0)$) {\footnotesize $\mathbf{6\times}$};

\begin{scope}[local bounding box=resgrid4,shift={($(res4.north)+(0,.3)$)}, every node/.append style={
		yslant=0,xslant=1},yslant=0,xslant=1
	]
	\fill[white,fill opacity=0.9] (-.4,-.16) rectangle (.4,.16);
	\draw[black,very thick] (-.4,-.16) rectangle (.4,.16);
\end{scope}

\node (res5) [above = .7 of resgrid4, draw, rounded corners] {\footnotesize ResNet block 5};
\node (res5_num) [circle,draw=black, fill=black, text=white, inner sep=0pt,minimum size=5pt] at ($(res5.west)-(0.2,0)$) {\footnotesize $\mathbf{3\times}$};

\begin{scope}[local bounding box=resgrid5,shift={($(res5.north)+(0,.3)$)}, every node/.append style={
		yslant=0,xslant=1},yslant=0,xslant=1
	]
	\fill[white,fill opacity=0.9] (-.25,-.1) rectangle (.25,.1);
	\draw[black,very thick] (-.25,-.1) rectangle (.25,.1);
\end{scope}

\begin{scope}[local bounding box=conv_resfpn,shift={($(resgrid5.east)+(1.2, 0)$)}, every node/.append style={
		yslant=.5,xslant=0,solid},yslant=.5,xslant=0,solid
	]
	\fill[white,fill opacity=0.9] (-.08,-.2) rectangle (.08,.2);
	\draw[black,very thick] (-.08,-.2) rectangle (.08,.2);
\end{scope}
\node [below right = .4 and .5 of conv_resfpn] {\footnotesize $1\times1$ conv};

\draw [-{stealth[length=3mm]}, rounded corners=3pt] (temp_vel_concat) -| (stem);
\draw [-{stealth[length=3mm]}] (stem) -- (resgrid1);
\draw [-{stealth[length=3mm]}] (resgrid1) -- (res2);
\draw [-{stealth[length=3mm]}] (res2) -- (resgrid2);
\draw [-{stealth[length=3mm]}] (resgrid2) -- (res3);
\draw [-{stealth[length=3mm]}] (res3) -- (resgrid3);
\draw [-{stealth[length=3mm]}] (resgrid3) -- (res4);
\draw [-{stealth[length=3mm]}] (res4) -- (resgrid4);
\draw [-{stealth[length=3mm]}] (resgrid4) -- (res5);
\draw [-{stealth[length=3mm]}] (res5) -- (resgrid5);
\draw [-{stealth[length=3mm]}] (resgrid5) -- (conv_resfpn);

\node (fpn4) [right = 2.5 of resgrid4, draw, rounded corners] {\footnotesize FPN block 4};
\begin{scope}[local bounding box=fpngrid4,shift={($(fpn4.south)-(0,.3)$)}, every node/.append style={
		yslant=0,xslant=1},yslant=0,xslant=1
	]
	\fill[white,fill opacity=0.9] (-.4,-.16) rectangle (.4,.16);
	\draw[black,very thick] (-.4,-.16) rectangle (.4,.16);
\end{scope}

\node (fpn3) [right = 2.5 of resgrid3, draw, rounded corners] {\footnotesize FPN block 3};
\begin{scope}[local bounding box=fpngrid3,shift={($(fpn3.south)-(0,.6)$)}, every node/.append style={
		yslant=0,xslant=1},yslant=0,xslant=1
	]
	\fill[white,fill opacity=0.9] (-.6,-.24) rectangle (.6,.24);
	\draw[black,very thick] (-.6,-.24) rectangle (.6,.24);
\end{scope}

\draw [-{stealth[length=3mm]}, rounded corners=3pt] (conv_resfpn) -| (fpn4);
\draw [-{stealth[length=3mm]}] (fpn4) -- (fpngrid4);
\draw [-{stealth[length=3mm]}] (fpngrid4) -- (fpn3);
\draw [-{stealth[length=3mm]}] (fpn3) -- (fpngrid3);

\draw [-{stealth[length=3mm]}] (resgrid3) -- (fpn3);
\draw [-{stealth[length=3mm]}] (resgrid4) -- (fpn4);


\node [draw, circle, below=.5 of fpngrid3.south, inner sep =1pt] (fmap_sum) {$+$};

\begin{scope}[local bounding box=head_conv0,shift={($(fmap_sum.east)+(.6, 0)$)}, every node/.append style={
		yslant=.5,xslant=0,solid},yslant=.5,xslant=0,solid
	]
	\fill[white,fill opacity=0.9] (-.16,-.4) rectangle (.16,.4);
	\draw[black,very thick] (-.16,-.4) rectangle (.16,.4);
\end{scope}
\begin{scope}[local bounding box=head_conv1,shift={($(head_conv0.east)+(+.3, 0)$)}, every node/.append style={
		yslant=.5,xslant=0,solid},yslant=.5,xslant=0,solid
	]
	\fill[white,fill opacity=0.9] (-.16,-.4) rectangle (.16,.4);
	\draw[black,very thick] (-.16,-.4) rectangle (.16,.4);
\end{scope}

\coordinate [right= .4 of head_conv1.east] (out_waypoint);

\begin{scope}[local bounding box=box,shift={($(out_waypoint)+(.8, 0)$)}, every node/.append style={
		yslant=.5,xslant=0,solid},yslant=.5,xslant=0,solid
	]
	\fill[white,fill opacity=0.9] (-.08,-.2) rectangle (.08,.2);
	\draw[black,very thick] (-.08,-.2) rectangle (.08,.2);
\end{scope}
\node[right=.3 of box.east] {\footnotesize box};

\begin{scope}[local bounding box=cls,shift={($(out_waypoint)+(.8, .8)$)}, every node/.append style={
		yslant=.5,xslant=0,solid},yslant=.5,xslant=0,solid
	]
	\fill[white,fill opacity=0.9] (-.08,-.2) rectangle (.08,.2);
	\draw[black,very thick] (-.08,-.2) rectangle (.08,.2);
\end{scope}
\node[right=.3 of cls.east, align=center, font=\footnotesize] {class\\score};

\begin{scope}[local bounding box=vxvy,shift={($(out_waypoint)+(.8, -.8)$)}, every node/.append style={
		yslant=.5,xslant=0,solid},yslant=.5,xslant=0,solid
	]
	\fill[white,fill opacity=0.9] (-.08,-.2) rectangle (.08,.2);
	\draw[black,very thick] (-.08,-.2) rectangle (.08,.2);
\end{scope}
\node[right=.3 of vxvy.east] {\footnotesize $v_x, v_y$};

\begin{scope}[local bounding box=conv0_bypass,shift={($(resgrid1.south east)+(+.4, -.4)$)}, every node/.append style={
		yslant=.5,xslant=0,solid},yslant=.5,xslant=0,solid
	]
	\fill[white,fill opacity=0.9] (-.08,-.2) rectangle (.08,.2);
	\draw[black,very thick] (-.08,-.2) rectangle (.08,.2);
\end{scope}
\begin{scope}[local bounding box=conv1_bypass,shift={($(conv0_bypass.east)+(.2, 0)$)}, every node/.append style={
		yslant=.5,xslant=0,solid},yslant=.5,xslant=0,solid
	]
	\fill[white,fill opacity=0.9] (-.08,-.2) rectangle (.08,.2);
	\draw[black,very thick] (-.08,-.2) rectangle (.08,.2);
\end{scope}
\node [above = .4 of conv0_bypass] {\footnotesize bypass conv};

\draw [-{stealth[length=3mm]}] (fpngrid3) -- (fmap_sum);
\draw [-{stealth[length=3mm]}, rounded corners=3pt] (vel_grid_split) |- (conv0_bypass);
\draw (conv0_bypass) -- (conv1_bypass);
\draw [-{stealth[length=3mm]}, rounded corners=3pt] (conv1_bypass) -| node [midway, right] {\footnotesize (c)} (fmap_sum);

\draw [-{stealth[length=3mm]}] (fmap_sum) -- (head_conv0);
\draw [-{stealth[length=3mm]}] (head_conv0) -- (head_conv1);
\draw (head_conv1) -- (out_waypoint);
\draw [-{stealth[length=3mm]}, rounded corners=3pt] (out_waypoint) |- (cls);
\draw [-{stealth[length=3mm]}] (out_waypoint) -- (box);
\draw [-{stealth[length=3mm]}, rounded corners=3pt] (out_waypoint) |- (vxvy);


\node (resnet_detail) [below left = 0.0 and 6.0 of res5, draw, dashed] {
	\begin{tikzpicture}
	\begin{scope}[local bounding box=res_main1, every node/.append style={
		yslant=0,xslant=1,solid},yslant=0,xslant=1,solid
	]
	\fill[white,fill opacity=0.9] (-.4,-.16) rectangle (.4,.16);
	\draw[black,very thick](-.4,-.16) rectangle (.4,.16);
	\end{scope}
	\node [below left = .35 and 1.3 of res_main1] {\footnotesize $1\times1$ conv, BN, ReLU};
	
	\begin{scope}[local bounding box=res_main2,shift={($(res_main1.north)+(0,.5)$)}, every node/.append style={
		yslant=0,xslant=1,solid},yslant=0,xslant=1,solid
	]
	\fill[white,fill opacity=0.9] (-.4,-.16) rectangle (.4,.16);
	\draw[black,very thick] (-.4,-.16) rectangle (.4,.16);
	\end{scope}
	\node [below left = .35 and 1.3 of res_main2] {\footnotesize $3\times3$ conv, BN, ReLU};
	
	\begin{scope}[local bounding box=res_main3,shift={($(res_main2.north)+(0,.5)$)}, every node/.append style={
		yslant=0,xslant=1,solid},yslant=0,xslant=1,solid
	]
	\fill[white,fill opacity=0.9] (-.4,-.16) rectangle (.4,.16);
	\draw[black,very thick] (-.4,-.16) rectangle (.4,.16);
	\end{scope}
	\node [below left = .35 and 1.3 of res_main3] {\footnotesize $1\times1$ conv, BN\hphantom{, ReLU}};
	
	\begin{scope}[local bounding box=res_bypass,shift={($(res_main2.east)+(1.0,0)$)}, every node/.append style={
		yslant=0,xslant=1,solid},yslant=0,xslant=1,solid
	]
	\fill[white,fill opacity=0.9] (-.4,-.16) rectangle (.4,.16);
	\draw[black,very thick] (-.4,-.16) rectangle (.4,.16);
	\end{scope}
	\node [below right = .35 and 1.1 of res_bypass] {\footnotesize ($1\times1$ conv, BN)};
	
	\node (sum) [circle,inner sep=1pt, draw, solid, above right = .5 and .8 of res_main3] {+};
	\coordinate [above= .3 of sum] (out);
	
	\coordinate [below right= .3 and .8 of res_main1] (inwaypoint);
	\coordinate [below = .3 of inwaypoint] (in);
	
	\draw [rounded corners=3pt, solid,-{stealth}] (in) -- (inwaypoint);
	\draw [rounded corners=3pt, solid,-{stealth}] (inwaypoint) -| (res_main1);
	\draw [solid,-{stealth}] (res_main1) -- (res_main2);
	\draw [solid,-{stealth}] (res_main2) -- (res_main3);
	\draw [rounded corners=3pt, solid,-{stealth}] (res_main3) |- (sum);
	\draw [rounded corners=3pt, solid,-{stealth}] (inwaypoint) -| (res_bypass);
	\draw [rounded corners=3pt, solid,-{stealth}] (res_bypass) |- (sum);
	\draw [solid,-{stealth}] (sum) -- (out);
	\end{tikzpicture}
};
\draw [dashed] (res5_num.north west) -- (resnet_detail.north east);
\draw [dashed] (res5_num.south west) -- (resnet_detail.south east);

\node (fpn_detail) [above right = 0.5 and 3.0 of fpn4, draw, dashed] {
	\begin{tikzpicture}
	\node (sum) [circle,inner sep=1pt, draw, solid] {+};
	
	\begin{scope}[local bounding box=transconv,shift={($(sum.north)+(0, .5)$)}, every node/.append style={
		yslant=0,xslant=1,solid},yslant=0,xslant=1,solid
	]
	\fill[white,fill opacity=0.9] (-.4,-.16) rectangle (.4,.16);
	\draw[black,very thick] (-.4,-.16) rectangle (.4,.16);
	\end{scope}
	\node [above left = .35 and .7 of transconv, text width= 7em] {\footnotesize $3\times3$\\[-1.5ex]transposed conv};
	
	\begin{scope}[local bounding box=conv_short,shift={($(sum.west)+(-.8, 0)$)}, every node/.append style={
		yslant=.5,xslant=0,solid},yslant=.5,xslant=0,solid
	]
	\fill[white,fill opacity=0.9] (-.16,-.4) rectangle (.16,.4);
	\draw[black,very thick] (-.16,-.4) rectangle (.16,.4);
	\end{scope}
	\node [below = .6 of conv_short] {\footnotesize $1\times1$ conv};
	
	\coordinate [left=1 of conv_short] (in_short);
	\coordinate [above=.8 of transconv] (in_conv);
	\coordinate [below=.8 of sum] (out);
	
	\draw [solid,-{stealth}] (in_short) -- (conv_short);
	\draw [solid,-{stealth}] (conv_short) -- (sum);
	\draw [solid,-{stealth}] (in_conv) -- (transconv);
	\draw [solid,-{stealth}] (in_short) -- (conv_short);
	\draw [solid,-{stealth}] (transconv) -- (sum);
	\draw [solid,-{stealth}] (sum) -- (out);
	
	\end{tikzpicture}
};
\draw [dashed] (fpn4.north east) -- (fpn_detail.north west);
\draw [dashed] (fpn4.south east) -- (fpn_detail.south west);

\end{tikzpicture}

%% file: figures/matching.tex
\begin{tikzpicture}[node distance=2mm and 3mm]

\def\pdfheight{1.35cm}

\node (detnet_t0) [rotate=90] {
	\begin{tikzpicture}[scale=0.7]
	\draw (-1.1,.8) -- (1.1,.8) -- (1.1,-.7) -- (0,-1.0) -- (-1.1,-.7) -- cycle;
	\begin{scope}[scale=.5, shift={(0,.5)}, every node/.append style={
		yslant=0.5,xslant=-1},yslant=0.5,xslant=-1
	]

	\draw[fill=white] ($(-.85,-.85)+(-1.5,-1.5)$) rectangle ($(.85,.85)+(-1.5,-1.5)$);
	\draw[black,very thick] ($(-.85,-.85)+(-1.5,-1.5)$) rectangle ($(.85,.85)+(-1.5,-1.5)$);
	
	\draw[fill=white] ($(-.7,-.7)+(-1.2,-1.2)$) rectangle ($(.7,.7)+(-1.2,-1.2)$);
	\draw[black,very thick] ($(-.7,-.7)+(-1.2,-1.2)$) rectangle ($(.7,.7)+(-1.2,-1.2)$);

	\draw[fill=white] ($(-.7,-.7)+(-0.9,-0.9)$) rectangle ($(.7,.7)+(-0.9,-0.9)$);
	\draw[black,very thick] ($(-.7,-.7)+(-0.9,-0.9)$) rectangle ($(.7,.7)+(-0.9,-0.9)$);
	
	\draw[fill=white] ($(-.85,-.85)+(-0.6,-0.6)$) rectangle ($(.85,.85)+(-0.6,-0.6)$);
	\draw[black,very thick] ($(-.85,-.85)+(-0.6,-0.6)$) rectangle ($(.85,.85)+(-0.6,-0.6)$);
	
	\draw[fill=white] ($(-1,-1)+(-0.3,-0.3)$) rectangle ($(1,1)+(-0.3,-0.3)$);
	\draw[black,very thick] ($(-1,-1)+(-0.3,-0.3)$) rectangle ($(1,1)+(-0.3,-0.3)$);

	\fill[white,fill opacity=0.9] (-1,-1) rectangle (1,1);
	
	\fill[red] (0.6,0.6) rectangle (0.4,0.4); 
	\fill[blue] (-0.2,0.2) rectangle (-0.4,0.0); 
	\fill[orange] (-0.4,-0.6) rectangle (-0.6,-0.8); 
	\fill[red!50!white] (-0.2,0.0) rectangle (0.0,0.2); 
	\fill[blue!50!white] (0.2,-0.8) rectangle (0.0,-1.0); 
	\fill[orange!50!white] (0.0,-0.2) rectangle (-0.2,0.0);  
	
	\draw[step=.2, black] (-1,-1) grid (1,1); 
	\draw[black,very thick] (-1,-1) rectangle (1,1);
	\end{scope}
	\end{tikzpicture}	
};

\node (detnet_t1) [below=1.5 of detnet_t0.north west, rotate=90] {
	\begin{tikzpicture}[scale=0.7]
	\draw (-1.1,.8) -- (1.1,.8) -- (1.1,-.7) -- (0,-1.0) -- (-1.1,-.7) -- cycle;
	\begin{scope}[scale=.5, shift={(0,.5)}, every node/.append style={
		yslant=0.5,xslant=-1},yslant=0.5,xslant=-1
	]

	\draw[fill=white] ($(-.85,-.85)+(-1.5,-1.5)$) rectangle ($(.85,.85)+(-1.5,-1.5)$);
	\draw[black,very thick] ($(-.85,-.85)+(-1.5,-1.5)$) rectangle ($(.85,.85)+(-1.5,-1.5)$);
	
	\draw[fill=white] ($(-.7,-.7)+(-1.2,-1.2)$) rectangle ($(.7,.7)+(-1.2,-1.2)$);
	\draw[black,very thick] ($(-.7,-.7)+(-1.2,-1.2)$) rectangle ($(.7,.7)+(-1.2,-1.2)$);

	\draw[fill=white] ($(-.7,-.7)+(-0.9,-0.9)$) rectangle ($(.7,.7)+(-0.9,-0.9)$);
	\draw[black,very thick] ($(-.7,-.7)+(-0.9,-0.9)$) rectangle ($(.7,.7)+(-0.9,-0.9)$);
	
	\draw[fill=white] ($(-.85,-.85)+(-0.6,-0.6)$) rectangle ($(.85,.85)+(-0.6,-0.6)$);
	\draw[black,very thick] ($(-.85,-.85)+(-0.6,-0.6)$) rectangle ($(.85,.85)+(-0.6,-0.6)$);
	
	\draw[fill=white] ($(-1,-1)+(-0.3,-0.3)$) rectangle ($(1,1)+(-0.3,-0.3)$);
	\draw[black,very thick] ($(-1,-1)+(-0.3,-0.3)$) rectangle ($(1,1)+(-0.3,-0.3)$);

	\fill[white,fill opacity=0.9] (-1,-1) rectangle (1,1);
	
	\fill[red] (0.6,0.6) rectangle (0.4,0.4); 
	\fill[blue] (0.0,0.0) rectangle (-0.2,-0.2); 
	\fill[orange] (-0.4,-0.6) rectangle (-0.6,-0.8); 
	\fill[red!50!white] (0.0,-0.2) rectangle (+0.2,0.0); 
	\fill[blue!50!white] (0.2,-0.8) rectangle (0.0,-1.0); 
	\fill[orange!50!white] (0.2,-0.4) rectangle (0.0,-0.2);  
	
	\draw[step=.2, black] (-1,-1) grid (1,1); 
	\draw[black,very thick] (-1,-1) rectangle (1,1);
	\end{scope}
	\end{tikzpicture}	
};

\node (pc1) [right=of detnet_t1.south, draw] {\includegraphics[height=\pdfheight]{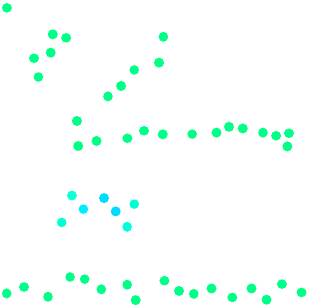}};
\node (pc1_obb0) [draw=blue, densely dotted, line width=1pt, minimum width=.42cm, minimum height=.1cm] at ($(pc1) +(-.25, -.26)$) {};
\draw [-{stealth[length=3mm]}, blue, line width=1pt] (pc1_obb0.east) -- node[near end, below right, blue, pos=.4] {\footnotesize$v$} ++(0.4,0);
\node (pc1_obb1) [draw=blue, densely dotted, line width=1pt, minimum width=.42cm, minimum height=.01cm, rotate=40] at ($(pc1) +(-.48, .44)$) {};
\draw [-{stealth[length=3mm]}, blue, line width=1pt] (pc1_obb1.east) -- ++(0.14,0.118) node[near end, right, pos=.] {\footnotesize$v$};

\node (pos_update) [right= of pc1.east, draw] {$+ v \cdot \Delta t$};

\node (pc1u) [right=of pos_update.east, draw] {\includegraphics[height=\pdfheight]{figures/pc_t1.pdf}};
\node (pc1u_obb0) [draw=blue, densely dotted, line width=1pt, minimum width=.42cm, minimum height=.1cm] at ($(pc1u) +(.45, -.26)$) {};
\node (pc1u_obb1) [draw=blue, densely dotted, line width=1pt, minimum width=.42cm, minimum height=.01cm, rotate=40] at ($(pc1u) +(-.45, .47)$) {};

\node (pc0) [draw] at (pc1u |- detnet_t0) {\includegraphics[height=\pdfheight]{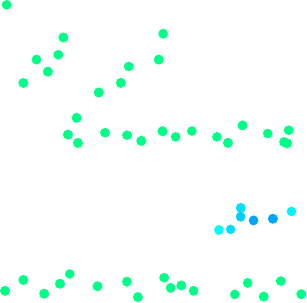}};
\node (pc0_obb0) [draw=blue, densely dotted, line width=1pt, minimum width=.42cm, minimum height=.1cm, rotate=15] at ($(pc0) +(.45, -.29)$) {};
\node (pc0_obb1) [draw=blue, densely dotted, line width=1pt, minimum width=.42cm, minimum height=.01cm, rotate=60] at ($(pc0) +(-.46, .38)$) {};

\node (labels) [above=1.65 of pc0.east, draw, minimum width=1.65cm, anchor=east] {\footnotesize box labels $@t_0$} ;

\node (helper) at ($(pc0)!0.5!(pc1u)$) {};
\node (lvel) [right=1.1 of helper, draw, minimum width={width("aaaa|")}] {$\mathcal{L}_{\text{vel}}$};
\draw [decorate,decoration={brace,amplitude=6pt}] ($(pc0.east)+(.0, -.2)$) -- node [midway] (lvel_helper) {} ($(pc1u.east)+(.0, .2)$);
\draw [-{stealth[length=3mm]}, rounded corners=3pt] ($(lvel_helper)+(.2,.0)$) -- (lvel.west);

\draw [decorate,decoration={brace,amplitude=6pt}] (labels.east) -- node [midway] (lcls_helper) {} ($(pc0.east)+(.0, .2)$);
\node (lcls) [draw, align=center, minimum width={width("aaaa|")}] at (lvel |- lcls_helper) {$\mathcal{L}_{\text{cls}}$,\\ $\mathcal{L}_{\text{box}}$};
\draw [-{stealth[length=3mm]}, rounded corners=3pt] ($(lcls_helper)+(.2,.0)$) -- (lcls.west);

\node [above=-.12 of detnet_t0.east] {\footnotesize $\mathcal{P}_{\text{det}} @t_0$};
\node [below=-.12 of detnet_t1.west] {\footnotesize $\mathcal{P}_{\text{vel}} @t_0 - \Delta t$};
\node [below=0.01 of pc1.south] {\footnotesize$\mathcal{B}_{\text{vel}} @t_0 - \Delta t$};
\node [below=0.01 of pc1u.south] {\footnotesize $\tilde{\mathcal{B}}_{\text{vel}} @t_0$};
\node [above=0.01 of pc0.north] {\footnotesize $\mathcal{B}_{\text{det}} @t_0$};
\node [below=0.01 of pos_update.south, align=center, font=\footnotesize] {position\\ update};
\node [align=center, font=\footnotesize] at ($(helper)-(1.1, 0.)$) {box\\matching};

\draw [-{stealth[length=3mm]}] (detnet_t0.south) -- (pc0);
\draw [-{stealth[length=3mm]}] (detnet_t1.south) -- (pc1);
\draw [-{stealth[length=3mm]}] (pc1.east) -- (pos_update.west);
\draw [-{stealth[length=3mm]}] (pos_update.east) -- (pc1u.west);


\draw [fill=black] (pc0_obb0.center) circle (.04);
\draw [fill=black] (pc0_obb1.center) circle (.04);
\draw [fill=black] (pc1u_obb0.center) circle (.04);
\draw [fill=black] (pc1u_obb1.center) circle (.04);
\draw [line width=1pt] (pc0_obb0.center) -- (pc1u_obb0.center);
\draw [line width=1pt] (pc0_obb1.center) -- (pc1u_obb1.center);

\end{tikzpicture}